\documentclass[9pt,twocolumn,twoside]{osajnl}

\usepackage{times}

\usepackage{amsmath}
\usepackage{amssymb}
\usepackage{multirow}
\usepackage{multirow}
\usepackage{mathtools}
\usepackage{units}
\usepackage{subfloat}
\usepackage{float}
\usepackage[]{units}
\usepackage{xcolor}
\usepackage{url}
\usepackage{bbm}
\usepackage{algorithm}
\usepackage{algpseudocode}
\usepackage{pbox}
\usepackage{placeins}
\usepackage{bm}
\usepackage{acronym}

\journal{ol} 

\setboolean{shortarticle}{false}

\title{Learning robust perceptive locomotion for quadrupedal robots in the wild}

\author[1,*]{Takahiro Miki}
\author[1]{Joonho Lee}
\author[2]{Jemin Hwangbo}
\author[1]{Lorenz Wellhausen}
\author[3]{Vladlen Koltun}
\author[1]{Marco Hutter}

\affil[1]{Robotic Systems Lab, ETH Zurich, Zurich, Switzerland}
\affil[2]{Robotics and Artificial Intelligence Lab, KAIST, Daejeon, Korea}
\affil[3]{Intelligent Systems Lab, Intel, Jackson, WY, USA.}
\affil[*]{Corresponding author: tamiki@ethz.ch}

\begin{abstract}
Legged robots that can operate autonomously in remote and hazardous environments will greatly increase opportunities for exploration into under-explored areas.
Exteroceptive perception is crucial for fast and energy-efficient locomotion: perceiving the terrain before making contact with it enables planning and adaptation of the gait ahead of time to maintain speed and stability. However, utilizing exteroceptive perception robustly for locomotion has remained a grand challenge in robotics. Snow, vegetation, and water visually appear as obstacles on which the robot cannot step~-- or are missing altogether due to high reflectance. Additionally, depth perception can degrade due to difficult lighting, dust, fog, reflective or transparent surfaces, sensor occlusion, and more. For this reason, the most robust and general solutions to legged locomotion to date rely solely on proprioception. This severely limits locomotion speed, because the robot has to physically feel out the terrain before adapting its gait accordingly. Here we present a robust and general solution to integrating exteroceptive and proprioceptive perception for legged locomotion. We leverage an attention-based recurrent encoder that integrates proprioceptive and exteroceptive input. The encoder is trained end-to-end and learns to seamlessly combine the different perception modalities without resorting to heuristics. The result is a legged locomotion controller with high robustness and speed.
The controller was tested in a variety of challenging natural and urban environments over multiple seasons and completed an hour-long hike in the Alps in the time recommended for human hikers.
\end{abstract}

\acrodef{RL}{Reinforcement Learning}
\acrodef{CPG}{central pattern generator}
\acrodef{MDP}{Markov Decision Process}
\acrodef{POMDP}{Partially Observable Markov Decision Process}
\acrodef{RNN}{Recurrent Neural Network}
\acrodef{TCN}{Temporal Convolutional Network}
\acrodef{PPO}{Proximal Policy Optimization}
\acrodef{MLP}{Multilayer Perceptron}
\acrodef{GRU}{Gated Recurrent Unit}
\acrodef{LSTM}{Long short-term Memory}
\acrodef{GPU}{graphics processing unit}

\begin{document}

\maketitle

\section{Introduction}
\begin{figure*}
   \centering
    \includegraphics[width=1.0\textwidth,height=0.8\textheight,keepaspectratio]{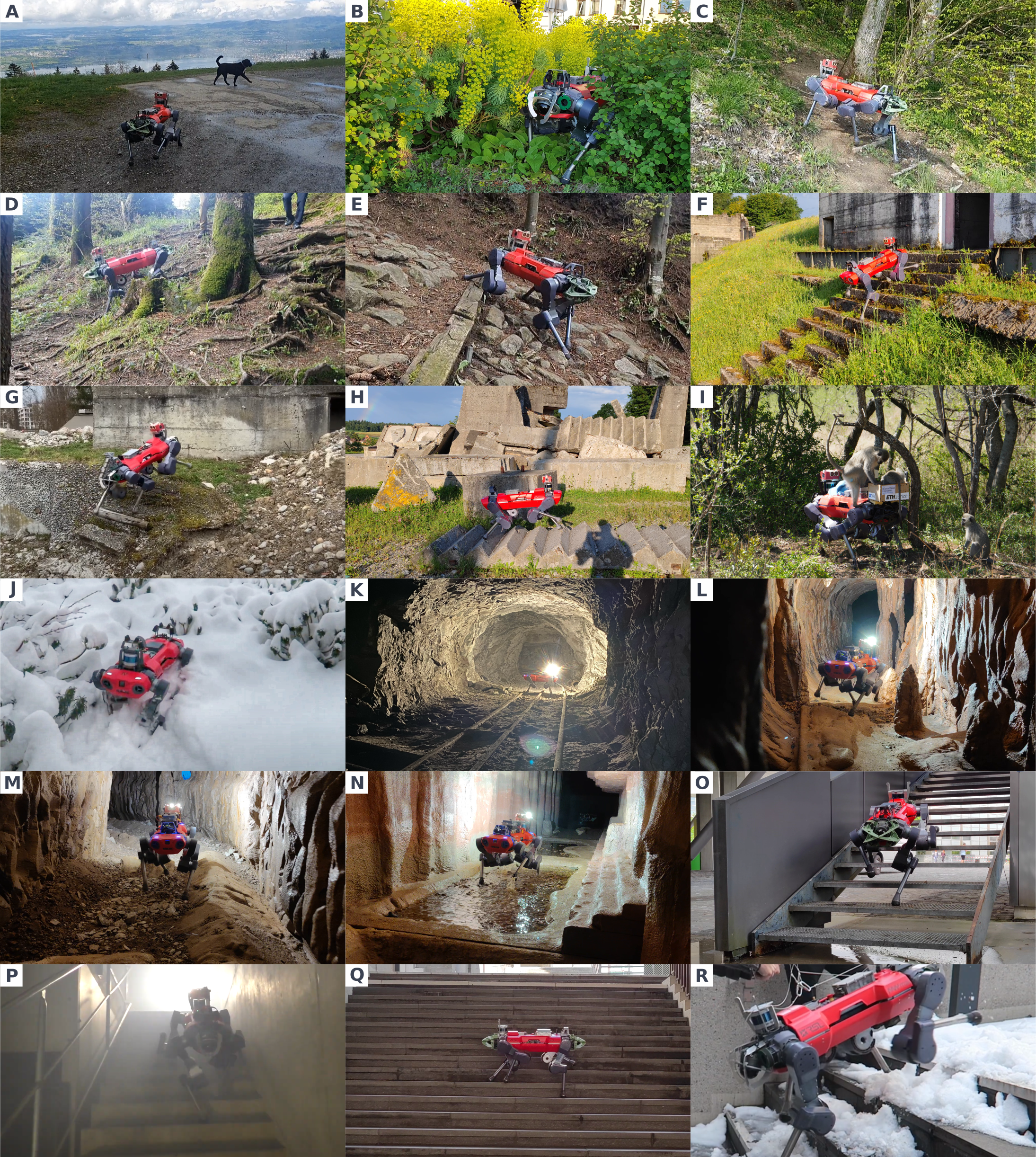}

    \caption{Robust locomotion in the wild. The presented locomotion controller was extensively tested in a variety of complex environments over multiple seasons. The controller overcame a whole spectrum of real-world challenges, often encountering them in combination. These include slippery surfaces, steep inclinations, complex terrain, and vegetation in natural environments. In search-and-rescue scenarios, the controller dealt with steep stairs, unknown payloads, and perception-degrading fog. Reflective surfaces, loose ground, low light, and water puddles were encountered in underground cave systems. Soft and slippery snow piled up in the winter. The controller traversed these environments with zero failures.}
    \label{fig:various_terrains}
\end{figure*}

Legged robots can carry out missions in challenging environments that are too far or too dangerous for humans, such as  hazardous  areas and the surfaces of other  planets.   Legs can  walk  over  challenging terrain with steep slopes, steps, and gaps that may impede wheeled or tracked vehicles of similar size.  There has been notable progress in legged robotics~\cite{raibert2008bigdog, katz2019mini, hwangbo2019learning, lee2020learning, park2021jumping}  and several commercial platforms are being deployed in the real world~\cite{spot,gehring2021anymal,agilityRobotics, unitree, ghost}.

However, until now, legged robots could not match the performance of animals in traversing challenging real-world terrain. %
Many legged animals such as humans and dogs can briskly walk or run in such environments by foreseeing the upcoming terrain and planning their footsteps based on visual information~\cite{matthis2018gaze}. 
Animals naturally combine proprioception and exteroception to adapt to highly irregular terrain shape and surface properties such as slipperiness or softness, even when visual perception is limited. 
Endowing legged robots with this ability is a grand challenge in robotics.

One of the biggest difficulties lies in reliable interpretation of incomplete and noisy perception for control.
Exteroceptive  information provided by onboard  sensors  is incomplete and often unreliable in real-world environments.  
Stereo camera based depth sensors, which most existing legged robots rely on~\cite{spot, unitree, anymal}, require texture to perform stereo matching and consequently struggle with low-texture surfaces or when parts of the image are under or overexposed.
Time of Flight (ToF) cameras often fail to perceive dark surfaces and become noisy under sunlight~\cite{fankhauser2015kinect}.
Generally, sensors which rely on light to infer distance are prone to producing artifacts on highly reflective surfaces, since the sensors assume that light travels in a straight path.
In addition, depth sensors by nature cannot distinguish soft unstable surfaces such as vegetation from rigid ones. 
An elevation map is commonly used to represent geometric terrain information extracted from depth sensor measurements~\cite{ye2003new, belter2011rough, fankhauser2014robot, fankhauser2018probabilistic}. 
It relies on the robot's estimated pose and is therefore affected by errors in this estimate. 
Other common sources of uncertainty in the map are occlusion or temporal inconsistency of the measurements due to dynamic objects. 
Most existing methods that rely on onboard terrain perception are still vulnerable to these failures.

 Conventional approaches assume that the terrain information and any uncertainties encoded in the map are reasonably accurate, and the focus shifts solely to generating the motion. Offline methods use a pre-scanned terrain map, compute a handcrafted cost function over the map, and optimize a trajectory which is replayed on the robot~\cite{zucker2010optimization, neuhaus2011comprehensive}. They assume perfect knowledge of the full terrain and robot states and plan complex motions with long planning times. Online methods generally employ a similar approach but use only onboard resources to construct a map and continuously replan trajectories during execution~\cite{kolter2009stereo, havoutis2013onboard, mastalli2017trajectory, belter2016adaptive, fankhauser2018robust}. Recently, faster locomotion has been achieved by reducing the planning time with heuristics~\cite{jenelten2020perceptive, kim2020vision, magana2019fast} or using Convolutional Neural Networks (CNN) to calculate foothold cost more efficiently~\cite{magana2019fast}.
Recently, a bipedal robot Atlas demonstrated parkour over complex obstacles~\cite{atlasparkour2021}. It leverages pre-planned motion reference and optimizes its motion online by utilizing onboard LiDAR sensor data.
Overall, the focus of all the approaches mentioned above is on picking footholds and generating trajectories given accurate terrain information. 
Some works~\cite{ye2003new, fankhauser2018probabilistic} represent the statistical uncertainty of the measurements in the map, but its use is limited to heuristically defined foot placement rules to avoid risky areas~\cite{fankhauser2018robust}. Such methods can only handle explicitly modeled uncertainties and are not robust to the variety of perception failures encountered in the wild.

Data-driven methods have recently been introduced in order to incorporate more complex dynamics without compromising real-time performance. 
Learning-based quadrupedal or bipedal locomotion for simulated characters has been achieved by using reinforcement learning (RL)~\cite{peng2016terrain, peng2017deeploco, 2018-TOG-deepMimic, xie2020allsteps} and realistic robot models were used in recent works~\cite{tsounis2020deepgait}. However, these works were only conducted in simulation.
 Recently, RL based locomotion controllers have been successfully transferred to physical robots~\cite{hwangbo2019learning, lee2020learning,  tan2018sim, RoboImitationPeng20,yang2020data, pmlr-v100-xie20a, siekmann2021blind, kumar2021rma, yang2020}.
Hwangbo et al.~\cite{hwangbo2019learning, lee2019robust} realized quadrupedal locomotion and recovery on flat ground with a physical robot by  using  learned  actuator  dynamics to facilitate simulation-to-reality (sim-to-real) transfer. 
Lee et al.~\cite{lee2020learning} extended this approach and enabled rough-terrain locomotion by simulating challenging terrain in a privileged training setup with an adaptive curriculum.
Peng et al.~\cite{RoboImitationPeng20} used imitation learning to transfer animal motion to a legged robot. However, these methods do not use any visual information.  

In order to  add exteroceptive information to locomotion learning, 
Gangapurwala et al.~\cite{gangapurwala2020rloc} combined a learning-based foothold planner and a model-based whole-body motion controller to transfer policies to the real world in a laboratory setting. 
Their applications are limited to rigid terrain with mostly flat surfaces and are still constrained in their deployment range. 
Their performance is tightly bound to the quality of the map, which often becomes unreliable in the field. 

In both model-based and learning-based approaches, the assumption of flawless map quality precludes the application of these methods in uncontrolled outdoor environments. Handling uncertainties in terrain perception remains an open problem. Existing controllers avoid catastrophic failures by simply refraining from using visual information in outdoor environments~\cite{katz2019mini, lee2020learning, siekmann2021blind} or by adding heuristically defined reflex rules~\cite{focchi2020heuristic,spotguide}.

Here we present a terrain-aware locomotion controller for quadrupedal robots that overcomes limitations of previous approaches and enables robust traversal of harsh natural terrain at unprecedented speeds (Movie 1).
At its core, the controller is based on a principled solution to incorporating exteroceptive perception into locomotion control. 

The key component is a recurrent encoder that combines proprioception and exteroception into an integrated belief state. The encoder is trained in simulation to capture ground-truth information about the terrain given exteroceptive observations that may be incomplete, biased, and noisy. The belief state encoder is trained end-to-end to integrate proprioceptive and exteroceptive data without resorting to heuristics. It learns to take advantage of the foresight afforded by exteroception to plan footholds and accelerate locomotion when exteroception is reliable, and can seamlessly fall back to robust proprioceptive locomotion when needed. The learned controller thus combines the best of both worlds: the speed and efficiency afforded by exteroception and the robustness of proprioception.

The controller is trained via privileged learning~\cite{chen2020learning}.
We first train a teacher policy via \ac{RL} with full access to privileged information in the form of the ground-truth state of the environment.
This privileged training enables the teacher policy to discover the optimal behavior given perfect knowledge of the terrain. 
We then train a student policy that only has access to information that is available in the field on the physical robot.
The student policy is built around our belief state encoder and trained via imitation learning.
The student policy learns to predict the teacher's optimal action given only partial and noisy observations of the environment.

Once the student policy is trained, we deploy it on the robot without any fine-tuning. The controller gets onboard sensor observations and a desired velocity command, and outputs each joint’s target position as the action. 
The robot perceives the environment by leveraging a robot-centric elevation map.
The elevation map serves as an abstraction layer between sensors and the locomotion controller, making our method independent of depth sensor choices. It works with no fine-tuning with different sensors, such as stereo cameras or LiDAR.
Since the policy was trained to handle significant noise, bias, and gaps in the elevation map, the robot can continue walking even when mapping fails or the sensors are physically broken.

The presented approach achieves substantial improvements over the state of the art~\cite{lee2020learning} in locomotion speed and obstacle traversability while maintaining exceptional robustness. 
Our key contribution is a method for combining multi-modal perception and demonstrating with extensive hardware experiments that the resulting control policy is robust against various exteroceptive failures.
 Handling exteroception failures has been a challenging problem in robotics. Our approach constitutes a general framework for robust deployment of complex autonomous machines in the wild.  

\section{Results}

\begin{figure*}
   \centering
    \includegraphics[width=\textwidth,height=0.9\textheight,keepaspectratio]{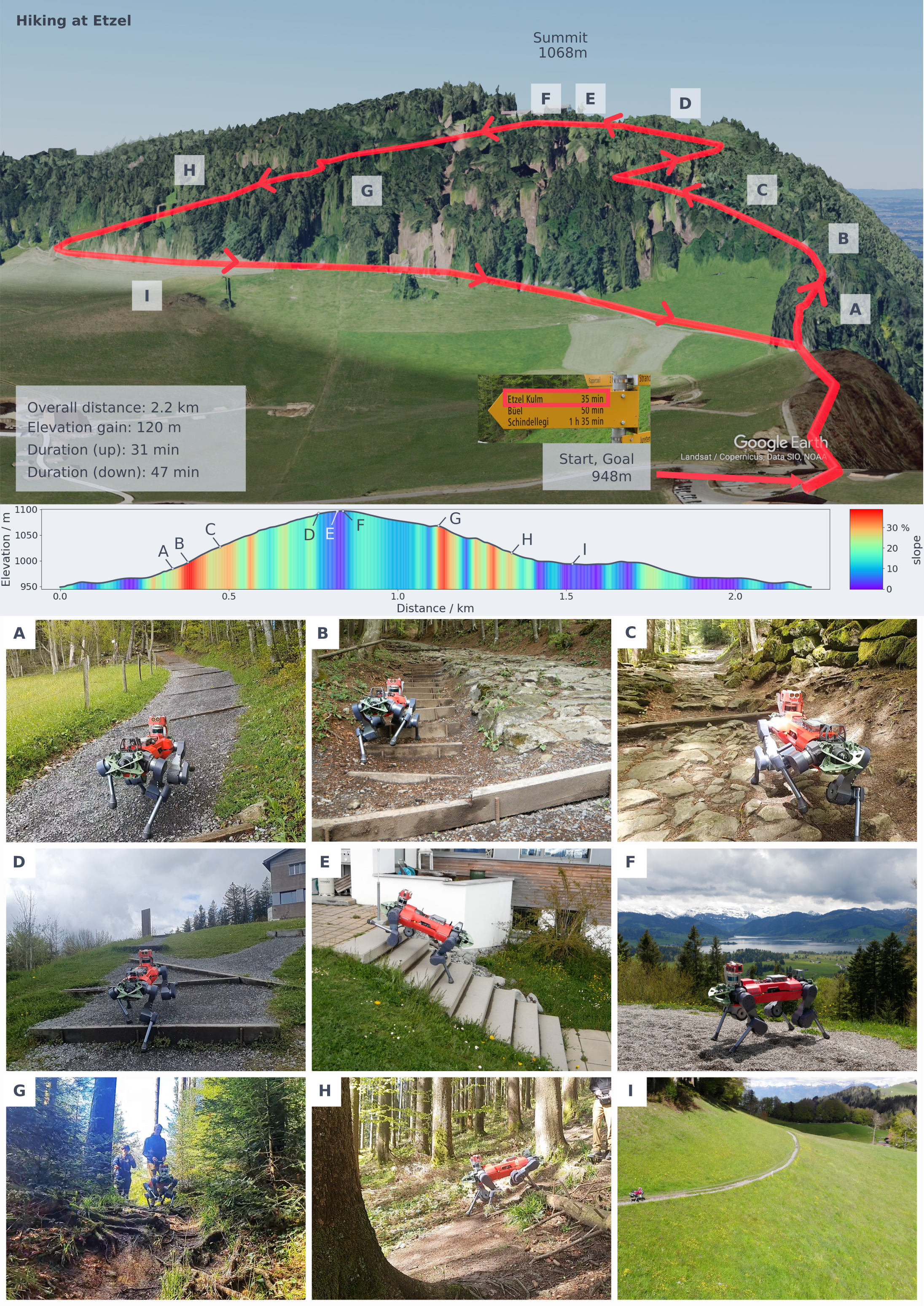}
    \caption{A hike on the Etzel mountain in Switzerland, completed by ANYmal with our locomotion controller. The 2.2km route~-- with 120m of elevation gain and inclinations up to 38\%~-- encompasses a variety of challenging terrain. ANYmal reached the summit faster than the human time indicated in the official signage, and finished the entire route in virtually the same time as given by a hiking guide~\cite{komoot}.}
    \label{fig:hiking}
\end{figure*}

\subsection*{Fast and robust locomotion in the wild}

We deployed our controller in a wide variety of terrain, as shown in Figure 1 and Movie 1. This includes alpine, forest, underground, and urban environments.
The controller was consistently robust and had zero falls during all deployments.
Because of the exteroceptive perception, the robot could anticipate the terrain and adapt its motion to achieve fast and smooth walking.
This was particularly notable for structures that require high foot clearance, such as stairs and large obstacles. The robot was able to leverage exteroceptive input to conquer terrain that was beyond the capabilities of prior work that did not utilize exteroception~\cite{lee2020learning}. 

ANYmal successfully traversed challenging natural environments with steep inclination, slippery surfaces, grass, and snow (Figure 1 A-J). The robot was robust in these conditions, even when occlusion and surface properties such as high reflectance impeded exteroception.
Our controller was also robustly deployed in underground environments with loose gravel, sand, dust, water, and limited illumination (Figure 1 K-N).

Urban environments also present important challenges (Figure 1 O-R). 
For traversing stairs, the state-of-the-art quadrupedal robot Spot from Boston Dynamics requires that a dedicated mode is engaged, and the robot must be properly oriented with respect to the stairs~\cite[p.~33]{spotguide}.
In contrast, our controller does not require any special mode for stairs, and can traverse stairs natively in any direction and any orientation, such as sideways, diagonally, and turning around on the stairway. %
See Movie S1 for demonstrations of smooth and robust stair traversal in arbitrary direction with our controller.

The controller was also robust to combinations of different challenges, as can be seen with snow on stairs in Figure 1R. 
Snow makes stairs slippery and yields incomplete and erroneous exteroceptive data. Depth sensors either fail due to the high reflectivity of snow, or estimate the surface profile to be on top of the snow, whereas the robot's legs sink below this level.
Foot slippage in snow can also cause large drift in the kinematic pose estimation~\cite{bloesch2013state}, making the map even more inconsistent.
Nevertheless, the controller remained consistently robust, with zero failures in this regime as well.

\subsection*{A hike in the Alps}

To further evaluate the robustness of our controller, we conducted a hiking experiment in which we tested if ANYmal could complete an hour-long hiking loop on the Etzel mountain in Switzerland. The hiking route was 2.2 km long, with an elevation gain of 120 m. Completing the trail required traversing steep inclinations, high steps, rocky surfaces, slippery ground, and tree roots (Figure 2). 
As seen in Movie 2, ANYmal completed the entire hike without any failure, stopping only to fix a detached shoe and swap batteries.
The robot was able to reach the summit in 31 minutes, which is faster than the expected human hiking duration indicated in the official signage (35 minutes as shown in Figure 2), and finished the entire path in 78 minutes~-- virtually the same duration suggested by a hiking planner (76 minutes), which rates the hike ``difficult''~\cite{komoot}.
The difficulty levels are chosen from ``easy'', ``moderate'', and ``difficult'', calculated by combining the required fitness level, sport type, and the
technical complexity~\cite{komoothelp}.

During the hike, the controller faced various challenges.
The ascending path reached inclinations of up to 38\% with rocky and wet surfaces (Figure 2 (B-C)).
On the descent through a forest, tree roots formed intricate obstacles and the ground proved very slippery (Figure 2 (G-H)).

Vegetation above the robot sometimes introduced severe artifacts into the estimated elevation map.
Despite all the challenges, the robot finished the hike without any human help and without a single fall.

\subsection*{Exteroceptive challenges}

\begin{figure*}
   \centering
    \includegraphics[width=1.0\textwidth]{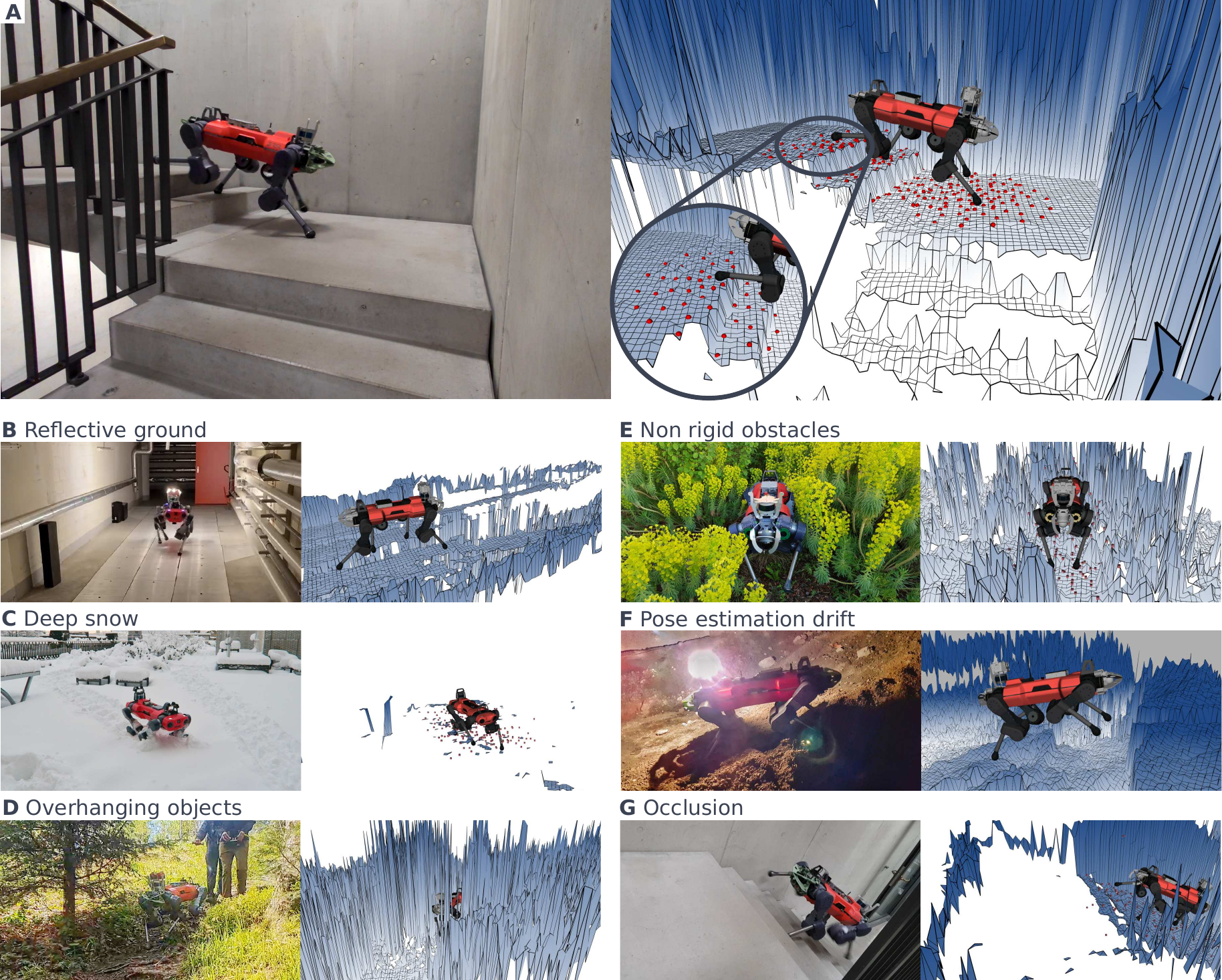}
    \caption{Our locomotion controller perceives the environment through height samples (red dots) from an elevation map (A). The controller is robust to many perception challenges commonly encountered in the field: missing map information due to sensing failure (B, C, G) and misleading map information due to non-rigid terrain (D, E) and pose estimation drift (F).}
    \label{fig:various_terrains}
\end{figure*}

In this section, we examine how the terrain was perceived by the robot in conditions that are challenging for exteroception.
The robot perceives the environment in the form of height samples from an elevation map constructed from point cloud input, as seen in Figure 3A.
We used LiDAR in some experiments (Figure 3D-G) and active stereo cameras in others (Figure 3B,C) to test the robustness of the controller to the sensing modality.

We encountered many circumstances in which exteroception provides incomplete or misleading input.
As shown in Figure 3 B-G, the estimated elevation map can unreliable due to sensing failures, limitations of the 2.5D height map representation, or viewpoint restrictions due to onboard sensing.

Since most depth sensors rely on light to infer distance, either through time-of-flight measurements or stereo disparity, they commonly struggle with reflective or translucent surfaces.Figure 3B shows such a sensing failure, where the reflective metal floor induced large depth outliers which appear as a trench in the elevation map. Figure 3C shows a sensing failure in the presence of snow. Since snow is highly reflective and has very little texture, stereo cameras could not infer depth, which lead to an empty map.

The 2.5D elevation map representation cannot accurately represent overhanging objects such as tree branches or low ceilings~\cite{fankhauser2018probabilistic}. These were integrated into the height field and were misrepresented as tall obstacles (Figure 3D). 
In addition, because the map cannot distinguish between rigid or soft materials, the map gave misleading information in soft vegetation or deep snow (Figure 3E). 

Slippery or deformable surfaces caused odometry drift because they violate the assumption of stable footholds, commonly adopted by kinematic pose estimators~\cite{bloesch2013state}. Since map construction relies on such pose estimation to register consecutive input point clouds, the map became inaccurate in such circumstances (Figure 3F).
Furthermore, since the sensors were only located on the robot itself, areas behind structures were occluded and not presented in the map, which was especially problematic during uphill walking (Figure~3G).

Overall, our controller could handle all of these challenging conditions gracefully, without a single failure. The belief state estimator was trained to assess the reliability of exteroceptive information and made use of it to the extent possible. When exteroceptive information was incomplete, noisy, or misleading, the controller could always gracefully degrade to proprioceptive locomotion, which was shown to be robust~\cite{lee2020learning}. The controller thus aims to achieve the best of both worlds: achieving fast predictive locomotion when exteroceptive information is informative, but seamlessly retaining the robustness of proprioceptive control when it is not.

\subsection*{Evaluating the contribution of exteroception}

\begin{figure*}
   \centering
     \includegraphics[width=1.0\textwidth]{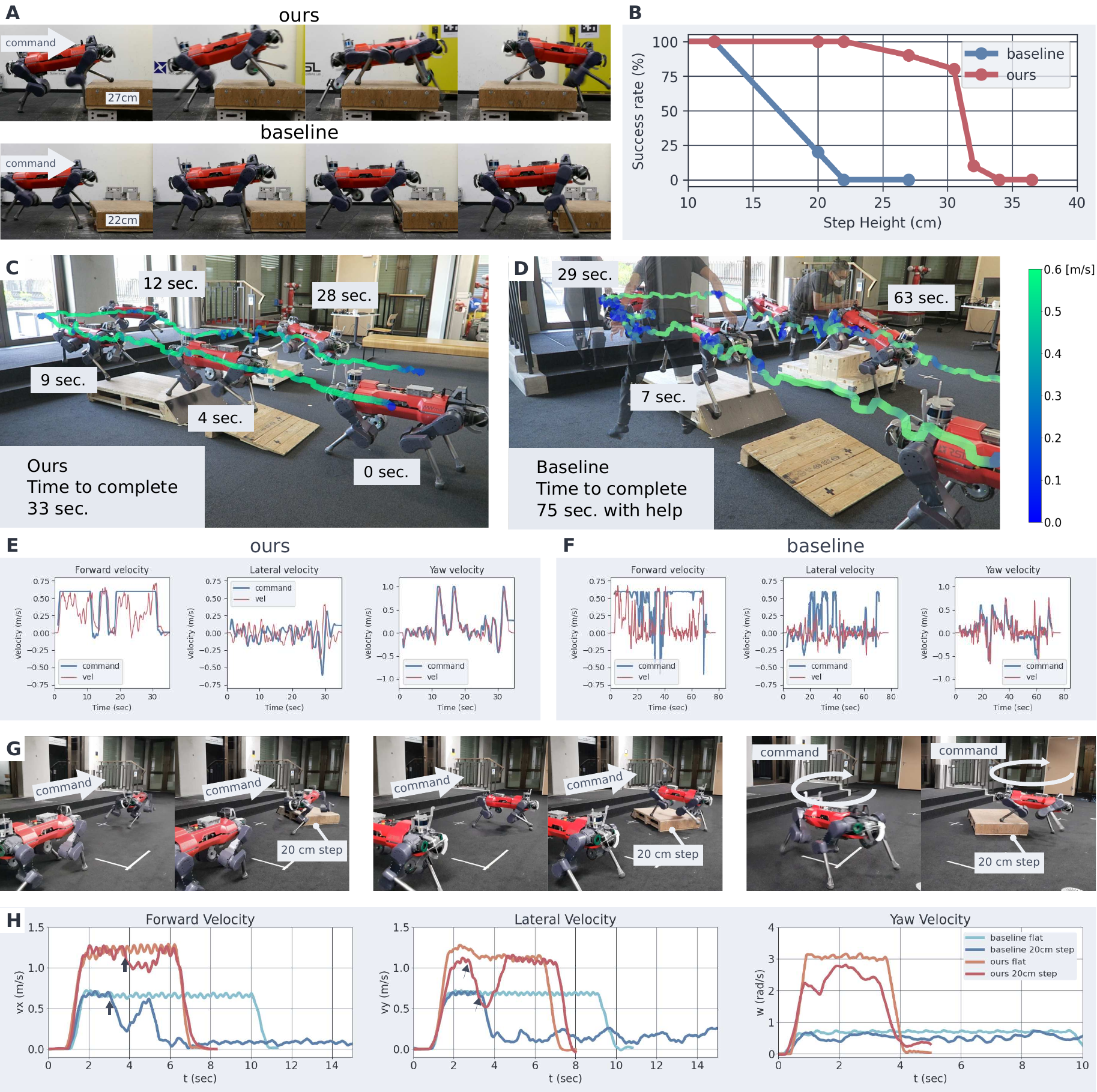}
    \label{fig:speed}
\end{figure*}

We conducted controlled experiments to quantitatively evaluate the contribution of exteroception. We compared our controller to a proprioceptive baseline~\cite{lee2020learning} that does not use exteroception.

First, we compared the success rate of overcoming fixed-height steps as shown in Figure 4A. Wooden steps of various height (from 12 cm to 36.5 cm) were placed ahead of the robot, which performed 10 trials to overcome each step with a fixed velocity command.
A trial was considered successful if the robot overcomes the step within 5 seconds.

The success rate of the proprioceptive baseline dropped at 20 cm step height when the front legs started frequently getting stuck at the step (Figure 4B). Even when the front legs successfully overcame the step, the hind legs often failed to fully step up.
In contrast, our controller reliably traversed steps of up to 30.5 cm in height. %
Since our controller could anticipate the step, it lifted its legs higher without making physical contact first, and leaned its body forward to let the hind leg swing over the step (Figure 4A).
Until this height, the dominating failure reason was the robot evading the step sideways instead of falling.
When approaching steps higher than 32 cm, our controller hesitated to walk forward because it learned that steps of such height are at or above the robot's physical limits and are likely to incur a high cost.

We also tested the two controllers in an obstacle course, as shown in Figure~4C,D. 
In this experiment, the robot was given a fixed path over the obstacles and tracked it using a pure pursuit controller~\cite{coulter1992implementation}.
The path traverses several types of obstacles~-- an inclined platform, a raised platform, stairs, and a pile of blocks.
The platforms are 20 cm high, the stairs are 17 cm high and 29 cm deep each, and the blocks are each 20 cm in both height and depth. Our controller followed the given path smoothly without any assistance, as shown in Figure~4C.
The exteroceptive perception provided advance information on the upcoming obstacles, allowing the controller to adjust the robot's motion before it made contact with the obstacles, facilitating fast and smooth motion through the obstacle course.
The baseline, on the other hand, failed to track the path without human assistance. During execution, it got stuck on all three obstacles and we had to lift and push the robot to continue the experiment (Figure~4D). 

In addition, we measured the maximum locomotion speed of both controllers over flat ground and in the presence of obstacles.
Figure 4E shows the experimental setup. We gave the controller a constant forward, lateral, or turning command and recorded the velocity on flat ground and over a 20 cm step. Note that the baseline controller only receives a directional command and learns to walk as fast as possible in the commanded direction~\cite{lee2020learning}.
Our controller walked at 1.2 m/s, while the baseline could only achieve 0.6 m/s on flat ground in both the forward and lateral directions. 
The difference became even more pronounced over the obstacle. 
Our controller could traverse the obstacle without any notable slow-down, while the baseline was stymied. 
The turning velocity showed the biggest difference between the baseline policy and ours. Our controller could turn at 3 rad/s while the baseline policy could only turn at 0.6 rad/s: a five-fold difference.

These results show clear gains by our controller over the proprioceptive baseline. Exteroception enabled our controller to traverse challenging environments more successfully and at higher speeds in comparison to pure proprioception.
Further quantitative performance evaluation is provided in the supplementary section S2.

\subsection*{Evaluating robustness with belief state visualization}

\begin{figure*}[htbp]
  \centering
   \includegraphics[width=\textwidth,height=0.75\textheight,keepaspectratio]{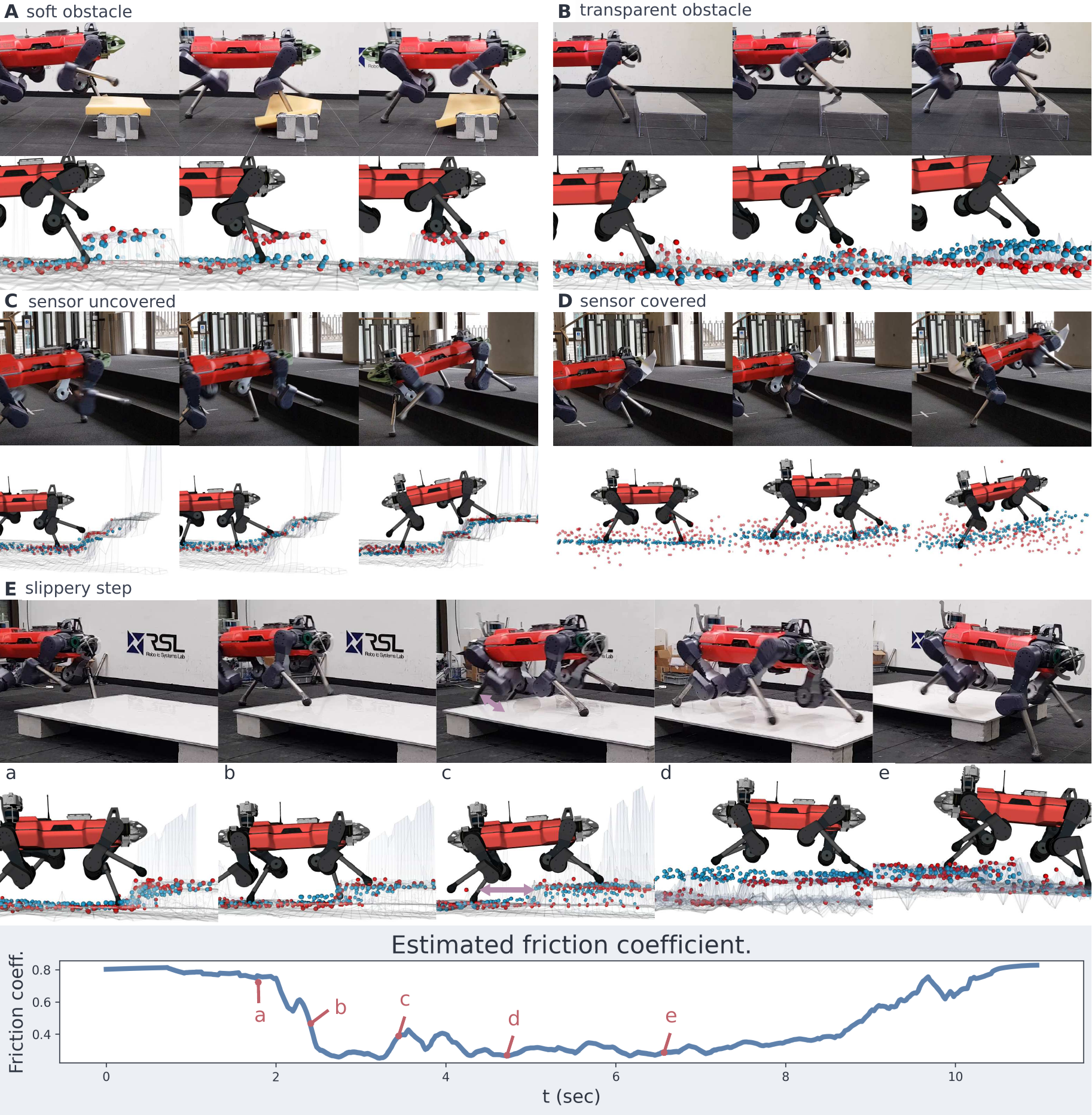}
    \caption{Internal belief state inspection during perceptive failure using a learned belief decoder.
    Red dots indicate height samples given as input to the policy. Blue dots show the controller's internal estimate of the terrain profile. 
    (A) After stepping on a soft obstacle that cannot support a foothold, the policy correctly revises its estimate of the terrain profile downwards. (B) A transparent obstacle is correctly incorporated into the terrain profile after contact is made. (C) With operational sensors, the robot swiftly and gracefully climbs the stairs, with no spurious contacts. (D) When the robot is blinded by covering the sensors, the policy can no longer anticipate the terrain but remains robust and successfully traverses the stairs. (E) When stepping onto a slippery platform, the policy identifies low friction and compensates for the induced pose estimation drift. The graph shows a decoded friction coefficient.}
    \label{fig:virtual_map}
\end{figure*}

To examine how our controller integrates proprioception and exteroception, we conducted a number of controlled experiments.
We tested with two types of obstacles that provide ambiguous or misleading exteroceptive input: an opaque foam obstacle that appears solid but cannot support a foothold, and a solid but transparent obstacle.
We placed each obstacle ahead of the robot and commanded the robot to walk forward at a constant velocity.

The sensors perceived the foam block as solid and the robot consequently prepared to step on it but could not achieve a stable foothold due to the deformation of the foam.
Figure~5A shows how the internal belief state (blue) was revised as the robot encounters the misleading obstacle: the controller initially trusted the exteroceptive input (red) but quickly revised its estimate of terrain height upon contact.
Once the correct belief had been formed, it was retained even after the foot left the ground, showing that the controller retains past information due to its recurrent structure.

The transparent obstacle is a block made of clear, acrylic plates, which are not accurately perceived by the onboard sensors (Figure~5B). 
The robot therefore walked as if it were on flat ground until it made contact with the step, at which point it revised its estimate of terrain profile upwards and changed its gait accordingly.

In the next experiment we simulated complete exteroception failure by physically covering the sensors, thus making them fully uninformative
(Figure 5C,D).
The robot was commanded to walk up and down two steps of stairs.
With an unobstructed sensor, the controller traversed the stairs gracefully, without any unintended contact with the stair risers, adjusting its footholds and body posture to step down the stairs softly.
When the sensors were covered, the map had no information and the controller received random noise as input.
In this condition, the robot made contact with the riser of the first stair, which could not be perceived in advance, revised its estimate of the terrain profile, adjusted its gait accordingly, and successfully climbed the stairs.
On the way down, the blinded robot made a hard landing with its front feet but kept its balance and stepped down softly with its hind legs.

Lastly, we tested locomotion over an elevated slippery surface (Figure~5E).
After the robot stepped onto the slippery platform, it detected the low friction and adapted its behavior to step faster and keep its balance.
The momentarily sliding feet violated the assumption of the kinematic pose estimator, which in turn destabilized the estimated elevation map and rendered exteroception uninformative during this time.
The controller seamlessly fell back on proprioception until the estimated elevation map stabilized and exteroception became informative again.

\section{Discussion}

We have presented a fast and robust quadrupedal locomotion controller for challenging terrain. The controller seamlessly integrates exteroceptive and proprioceptive input. Exteroceptive perception enables the robot to traverse the environment quickly and gracefully by anticipating the terrain and adapting its gait accordingly before contact is made. When exteroceptive perception is misleading, incomplete, or missing altogether, the controller smoothly transitions to proprioceptive locomotion. The controller remains robust in all conditions, including when the robot is effectively blind. The integration of exteroceptive and proprioceptive inputs is learned end-to-end and does not require any hand-coded rules or heuristics. The result is the first rough-terrain legged locomotion controller that combines the speed and grace of vision-based locomotion with the high robustness of proprioception.

This combination of speed and high robustness has been validated through controlled experiments and extensive deployments in the wild, including an hour-long hiking route in the Alps that is rated ``difficult"\cite{komoot}. 
The entire route was completed by the robot without human assistance (other than reattaching a detached shoe and swapping the batteries), in the recommended time for completion of this route by human hikers.

Our work expands the operational domain of legged robots and opens up new frontiers in autonomous navigation. Navigation planners no longer need to identify ground type or to switch modes during autonomous operation. Our controller was used as the default controller in the DARPA Subterranean Challenge missions of team Cerberus~\cite{tranzatto2021cerberus, cerberus} which has won the first prize in the finals~\cite{subtresult}.
In this challenge, our controller drove ANYmals to operate autonomously over extended periods of time in underground environments with rough terrain, obstructions, and degraded sensing in the presence of dust, fog, water, and smoke~\cite{subt}.
Our controller played a crucial role as it enabled four ANYmals to explore over 1700m in all three types of courses -- tunnel, urban, and cave -- without a single fall.

\subsection*{Possible extensions}

Future work could explicitly utilize the uncertainty information in the belief state. 
Currently, the policy uses uncertainty only implicitly to estimate the terrain. 
For example, in front of narrow cliff or a stepping stone, the elevation map does not provide sufficient information due to occlusion. Therefore, the policy assumes a continuous surface and, as a result, the robot might step off and fall.
Explicitly estimating uncertainty may allow the policy to become more careful when exteroceptive input is unreliable, for example using its foot to probe the ground if it is unsure about it. 
In addition, our current implementation obtains perceptual information through an intermediate state in the form of an elevation map, rather than directly ingesting raw sensor data. 
This has the advantage that the model is independent of the specific exteroceptive sensors. (We use LiDAR and stereo cameras in different deployments, with no retraining or fine-tuning.)
However, the elevation map representation omits detail that may be present in the raw sensory input and may provide additional information concerning material and texture. 
Furthermore, our elevation map construction relies on a classical pose estimation module that is not trained jointly with the rest of the system.
Appropriately folding the processing of raw sensory input into the network may further enhance the speed and robustness of the controller. In addition, an occlusion model could be learned, such that the policy understands that there's an occlusion behind the cliff and avoids stepping off it.
Another limitation is the inability to complete locomotion tasks which would require maneuvers very different from normal walking, for example recovering from a leg stuck in narrow holes or climbing onto high ledges.

\section{Materials and Methods}
\subsection*{Overview}

\begin{figure*}
   \centering
    \includegraphics[height=0.95\textheight]{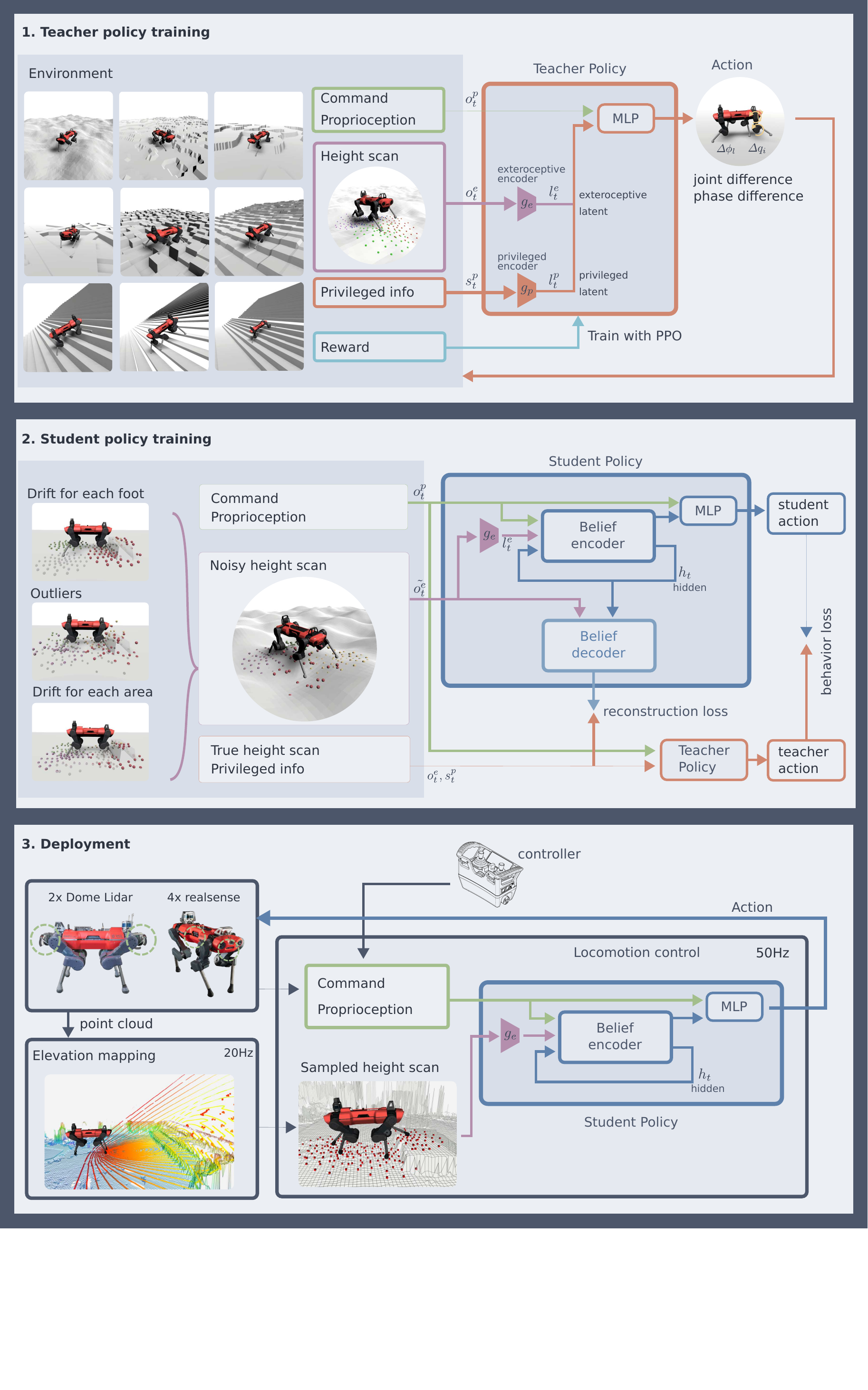}
    \caption{Overview of the training methods and deployment. We first train a teacher policy with access to privileged simulation data using reinforcement learning (RL). This teacher policy is then distilled into a student policy, which is trained to imitate the teacher's actions and to reconstruct the ground-truth environment state from noisy observations. We deploy the student policy zero-shot on real hardware using height samples from a robot-centric elevation map.}
    \label{fig:overview}
\end{figure*}

We train a neural network policy in simulation and then perform zero-shot sim-to-real transfer.
Our method consists of three stages, illustrated in Figure 6.

First, a teacher policy is trained with \ac{RL} to follow a random target velocity over randomly generated terrain with random disturbances. The policy has access to privileged information such as noiseless terrain measurements, ground friction, and the disturbances that were introduced.

In the second stage, a student policy is trained to reproduce the teacher policy's actions without using this privileged information. 
The student policy constructs a belief state to capture unobserved information using a recurrent encoder and outputs an action based on this belief state.
During training, we leverage two losses: a behavior cloning loss and a reconstruction loss.
The behavior cloning loss aims to imitate the teacher policy. The reconstruction loss encourages the encoder to produce an informative internal representation.

Lastly, we transfer the learned student policy to the physical robot and deploy it in the real world with onboard sensors. The robot constructs an elevation map by integrating depth data from onboard sensors, and samples height readings from the constructed elevation map to form the exteroceptive input to the policy. This exteroceptive input is combined with proprioceptive sensory data and is given to the neural network, which produces actuator commands.

\subsection*{Problem formulation}
We formulate our control problem in discrete time dynamics, where the environment is fully defined by the state $s_t$ at time step $t$.
The policy performs an action $a_t$ and observes the environment via $o_t$ which comes from an observation model $\mathcal{O}(o_t|s_t, a_t)$. 
Then, the environment moves to the next state $s_{t+1}$ with transition probability~$P(s_{t+1}|s_t, a_t)$ and returns a reward $r_{t+1}$.

When all states are observable such that $o_t = s_t$, this can be considered a \ac{MDP}.
When there is unobservable information, however, such as external forces or full terrain information in our case, the dynamics are modeled as a \ac{POMDP}.

The \ac{RL} objective is to find a policy $\pi^*$ that maximizes the expected discounted reward over the future trajectory, such that 
\begin{equation*}
    \pi^* = \mathop{\mathrm{argmax}}_a \mathop{\mathbb{E}}[\sum_{t=0}^\infty \gamma^tr_t].
\end{equation*} 

A number of \ac{RL} algorithms have been developed to solve fully-observable MDPs and are readily available to be used for training.
However, the case of POMDPs is more challenging since the state is not fully observable.
This is often overcome by constructing a belief state $b_t$ from a history of observations $\{o_0, \cdots,  o_t\}$ in an attempt to capture the full state.
In deep reinforcement learning, this is frequently done by stacking a sequence of previous observations~\cite{mnih2013playing} or by using architectures which can compress past information such as \acp{RNN}~\cite{zhu2017improving, vinyals2019grandmaster} or \acp{TCN}~\cite{BaiTCN2018,lee2020learning}.

Training a complex neural network policy that handles sequential data naively from scratch can be time-consuming~\cite{lee2020learning}.
Therefore we use privileged learning~\cite{chen2020learning}, in which we first train a teacher policy with privileged information, and then distill the teacher policy into a student policy via supervised learning.

\subsection*{Training environment}
We use RaiSim~\cite{raisim} as our simulator to build the training environment. There, we simulate multiple ANYmal-C robots on randomly generated rough terrain in parallel with an integrated actuator model~\cite{hwangbo2019learning} to close the reality gap.

\subsubsection*{Terrain}
We define parameterized terrain as shown in Figure 6.1. The terrain is modeled as a height map; further details are provided in supplementary section S4.

In addition to terrains composed of a variety of slopes and steps, we modelled four different types of stairs in the training environment; standard, open, ledged, and random.
We use boxes to form the stairs, because stair risers modeled by a height map are not perfectly vertical; we observed that the policy exploited these non-vertical edges in simulation, resulting in poor sim-to-real transfer.

\subsubsection*{Domain randomization}
We randomize the masses of the robot's body and legs, the initial joint position and velocity, and the initial body orientation and velocity in each episode.
In addition, external force and torque are applied to the body of the robot and the friction coefficients of the feet are occasionally set to a low value to introduce slippage.

\subsubsection*{Termination}
We terminate a training episode and start a new one when the robot reaches an undesirable state.
Termination criteria are: body collision with the ground, large body tilt, and exceeding the joint torque limit of the actuators.
These criteria help shape the motion and obtain constraint-satisfying behaviors.

\subsection*{Teacher policy training}
In the first stage of training we aim to find an optimal reference control policy which has access to perfect, privileged information and enables ANYmal to follow a desired command velocity over randomly generated terrain.
The desired command is generated randomly as a vector $\bm{v}_{des} \in \mathbb{R}^3 = (v_x, v_y, w)$, where $v_x, v_y$ represents the longitudinal and lateral velocity and $w$ represents the yaw velocity, all in the robot's body frame. 

We used \ac{PPO}~\cite{schulman2017proximal} to train the teacher policy.
The teacher is modeled as a Gaussian policy, $a_t \sim \mathcal{N}(\pi_{\theta}(o_t = s_t), \sigma I)$, where $\pi_{\theta}$ is implemented by a multilayer perceptron (MLP) parameterized by $\theta$, and $\sigma$ represents the variance for each action.

\subsubsection*{Observation and Action}
The teacher observation is defined as $o^{teacher}_t = (o_t^p, o_t^e, s_t^p)$, where $o_t^p$ refers to the proprioceptive observation, $o_t^e$ the exteroceptive observation, and $s_t^p$ the privileged state. $o_t^p$ contains the body velocity, orientation, joint position and velocity history, action history, and each leg's phase. $o_t^e$ is a vector of height samples around each foot with five different radii.
The privileged state $s_t^p$ includes contact states, contact forces, contact normals, friction coefficient, thigh and shank contact states, external forces and torques applied to the body, and swing phase duration.

Our action space is inspired by central pattern generators (CPGs)~\cite{lee2020learning}. Each leg $l = \{1,2,3,4\}$ keeps a phase variable $\phi_l$ and defines a nominal trajectory based on the phase. The nominal trajectory is a stepping motion of the foot tip and we calculate the nominal joint target $q_i(\phi_l)$ for each joint actuator $i = \{1, \cdots, 12\}$ using inverse kinematics.  
The action from the policy is the phase difference $\varDelta \phi_l$ and the residual joint position target $\varDelta q_i$.
More details of the observation and action space are in supplementary section S5.

\subsubsection*{Policy architecture}
We model the teacher policy $\pi_{\theta}$ as an MLP.
It consists of three MLP components: exteroceptive encoder, privileged encoder, and the main network, as shown in Figure 6.
The exteroceptive encoder $g_e$ receives $o_t^e$ and outputs a smaller latent representation $l_t^e$:
\begin{equation*}
    l_t^e = g_e(o_t^e)
\end{equation*}
The privileged encoder $g_{p}$ receives the privileged state $s_t^p$ and outputs a latent representation $l_t^{priv}$:
\begin{equation*}
    l_t^{priv} = g_{p}(s_t^p)
\end{equation*}
These encoders compress each input to a more compact representations and facilitate reuse of some of the teacher policy components by the student policy.
More details on each layer are in supplementary section S6

\subsubsection*{Rewards}
We define a positive reward for following the command velocity and a negative reward for violating some imposed constraints.
The command-following reward is defined as follows:
\begin{equation}
    r_{command} = 
    \begin{cases}
    1.0, & \text{if } \bm{v}_{des} \cdot \bm{v} > |\bm{v}_{des}|\\
    \exp(-(\bm{v}_{des} \cdot \bm{v} - |\bm{v}_{des}|)^2),& \text{otherwise}
     \end{cases}
\end{equation}
where $\bm{v}_{des} \in \mathbb{R}^2$ is the desired horizontal velocity and $\bm{v}\in \mathbb{R}^2$ is the current horizontal body velocity with respect to the body frame.
The same reward is applied to the yaw command as well.
We penalize the velocity component orthogonal to the desired velocity as well as the 
body velocity around roll, pitch, and yaw.
Additionally, we use shaping rewards for body orientation, joint torque, joint velocity, joint acceleration, foot slippage as well as
shank and knee collision.

Body orientation reward was used to avoid strange posture of the body.
Joint related reward terms were used to avoid overly aggressive motion.
Foot slippage and collision reward terms were used to avoid them.
We tuned the reward terms by looking at the policy's behavior in simulation.
In addition to the traversal performance, we checked the smoothness of the locomotion.
All reward terms are specified in supplementary section S7.

\subsubsection*{Curriculum}

We use two curricula to ramp up difficulty as the policy's performance improves. One curriculum adjusts terrain difficulty using an adaptive method~\cite{lee2020learning} and the other changes elements such as reward or applied disturbances using a logistic function~\cite{hwangbo2019learning}.

For the terrain curriculum, a particle filter updates the terrain parameters such that they remain challenging but achievable at any point during policy training~\cite{lee2020learning}.

The second curriculum multiplies the magnitude of domain randomization and some reward terms (joint velocity, joint acceleration, orientation, slip, thigh and shank contact) by a factor that is monotonically increasing and asymptotically trending to 1:
\begin{equation*}
    c_{k + 1} = (c_k)^d ,
\end{equation*}
where $c_{k}$ is the curriculum factor at the $k$th iteration and $0<d<1$ is the convergence rate.

\subsection*{Student policy training}
After we train a teacher policy that can traverse various terrain with the help of privileged information, we distill it into a student policy that only has access to information that is available on the real robot. We use the same training environment as for the teacher policy, but add additional noise to the student height sample observation: $o^{student}_{t} = (o_t^p, n(o_t^e))$, where $n(o_t^e)$ is a noise model applied to the height sample input. 
The noise model simulates different failure cases of exteroception frequently encountered during field deployment and is detailed below.

When there is a large noise in the exteroception, it becomes unobservable, thus the dynamics is considered to be POMDP. In addition, the privileged states are not observable due to the lack of sensors to directly measure. Therefore, the policy needs to consider the sequential correlation to estimate the unobservable states. We propose to use a recurrent belief state encoder to combine sequences of both exteroception and proprioception to estimate the unobservable states as a belief state.

The student policy consists of a recurrent belief state encoder and an MLP, as shown in Figure 6.2. 
We denote the hidden state of the recurrent network by $h_t$.
The belief state encoder takes $o^{student}_t$ and $h_t$ as input and outputs a latent vector $b_t$, which we refer to as the belief state. The goal is to match the belief state $b_t$ with the feature vector $(l_t^e, l_t^{priv})$ of the teacher policy which encodes all locomotion-relevant information.
We then pass $o_t^p$ and $b_t$ to the MLP which computes the output action.

The MLP structure remains the same as for the teacher policy, such that we can reuse the learned weights of the teacher policy to initialize the student network and speed up training.

Training is performed in supervised fashion by minimizing two losses: a behavior cloning loss and a reconstruction loss.
The behavior cloning loss is defined as the squared distance between the student action and the teacher action given the same state and command.
The reconstruction loss is the squared distance between the noiseless height sample and privileged information $(o_t^e, s_t^p)$ and their reconstruction from the belief state.
We generate samples by rolling out the student policy to increase robustness \cite{ross2011reduction, pmlr-v89-czarnecki19a}.

\subsubsection*{Height sample randomization}

\begin{figure*}[htbp]
   \centering
    \includegraphics[width=1.0\textwidth]{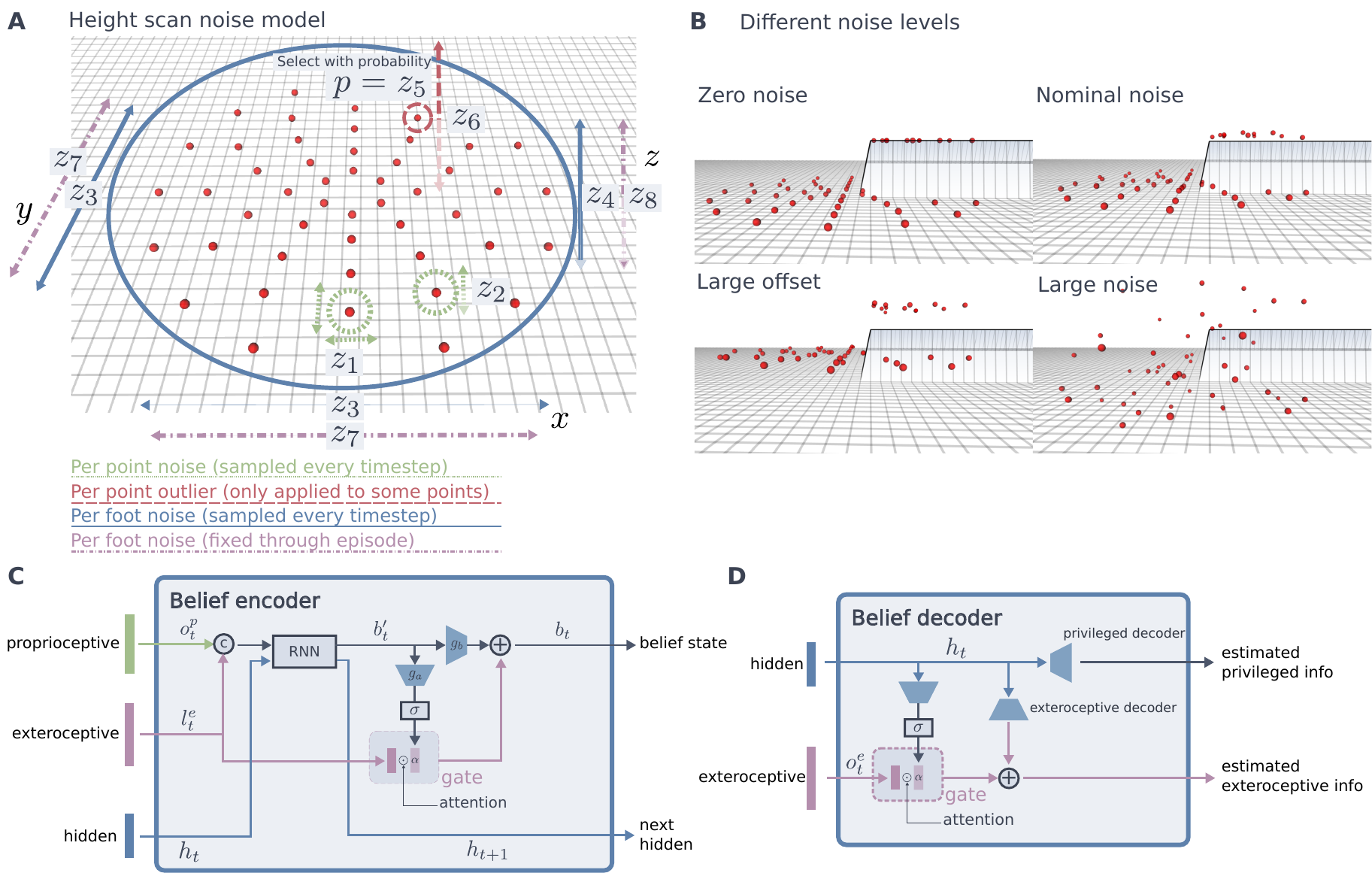}
    \caption{Details of robust terrain perception components. (A)~During student training, random noise is added to the height samples. The noise is sampled from a Gaussian distribution $\mathcal{N}(0,z^l \in \mathbb{R}^{8})$, where each $z^l_i$ controls a different noise component $i$ per leg $l$. 
    (B)~We use multiple noise configurations $z$ to simulate different operating conditions. ``Zero noise" is applied during teacher training, while ``nominal noise" represents normal mapping conditions during student training. ``Large offset" noise simulates large map offsets due to pose estimation drift or deformable terrain surfaces. ``Large noise" simulates a complete lack of terrain information due to occlusion or sensor failure.
    (C)~The student policy belief encoder incorporates a recurrent core and an attentional gate that integrates the proprioceptive and exteroceptive modalities. The gate explicitly controls which aspects of exteroceptive data to pass through. (D)~The belief decoder has a gate for reconstructing the exteroceptive data. It is only used during training and for introspection into the belief state. }
    \label{fig:priv_estimator}
\end{figure*}

During student training, we inject random noise into the height samples using a parameterized noise model $n(\Tilde{o_t^e}|o_t^e, z)$, $z \in \mathbb{R}^{8 \times 4}$.
We apply two different types of measurement noise when sampling the heights, as shown in Figure~7A: 
\begin{enumerate}
    \item Shifting scan points laterally.
    \item Perturbing the height values.
\end{enumerate}
Each noise value is sampled from a Gaussian distribution, and the noise parameter $z$ defines the variance. 
Both types of noise are applied in three different scopes, all with their own noise variance: per scan point, per foot, and per episode.
The noise values per scan point and per foot are resampled at every time step while the episodic noise remains constant for all scan points.

Additionally, we define three mapping conditions with associated noise parameters $z$ to simulate changing map quality and error sources, as shown in Figure 7B.
\begin{enumerate}
    \item Nominal noise assuming good map quality during regular operation.
    \item Large offsets through high per-foot noise to simulate map offsets due to pose estimation drift or deformable terrain.
    \item Large noise magnitude for each scan point to simulate complete lack of terrain information due to occlusion or mapping failure.
\end{enumerate}
These three mapping conditions are selected at the beginning of each training episode in a ratio of 60\%, 30\%, and 10\%.

Finally, we divide each training terrain into cells and add an additional offset to the height sample, depending on which cell it was sampled from. 
This simulates transitions between areas with different terrain characteristics, such as vegetation and deep snow.
The parameter vector $z$ is also part of a learning curriculum and its magnitude increases linearly with training duration.

The height sample representation is specified in more detail in supplementary section S8.

\subsubsection*{Belief state encoder}
The recurrent belief state encoder encodes states that are not directly observable.
To integrate proprioceptive and exteroceptive data, we introduce a gated encoder as shown in Figure 7C, inspired by gated \ac{RNN} models~\cite{cho2014learning, hochreiter1997long} and multimodal information fusion~\cite{anzai2020deep, kim2018robust, arevalo2017gated}.

The encoder learns an adaptive gating factor %
that controls how much exteroceptive information to pass through.
First, proprioception $o_t^p$, exteroceptive features from noisy observations $l_t^e = g_e(\Tilde{o_t^e})$, and hidden state $s_t$ are encoded by the \ac{RNN} module into the intermediate belief state $b_t'$.
Then, the attention vector $\alpha$ is computed from $b_t'$. It controls how much exteroceptive information enters the final belief state $b_t$:
\begin{eqnarray*}
    b_t', h_{t+1} &=& {\rm RNN} (o_t^p, l_t^e, h_t) \\
    \alpha &=& \sigma(g_{a}(b_t')) \\
    b_t &=& g_{b}(b_t') + l_t^e \odot  \alpha 
\end{eqnarray*}
Here, $g_{a}$ and $g_{b}$ are fully-connected neural networks and $\sigma(\cdot)$ is the sigmoid function.

The same gate is used in the decoder, where it is used to reconstruct the privileged information and the height samples (Figure~7D). This is used to calculate a reconstruction loss that encourages the belief state to capture veridical information about the environment.

We use the \ac{GRU}~\cite{cho2014learning} as our \ac{RNN} architecture.
The evaluation of the effectiveness of gate structure is presented in supplementary section S9.

\subsection*{Deployment}
We deployed our controller on the ANYmal C robot with two different sensor configurations, either using two Robosense Bpearl~\cite{bpearl} dome Lidar sensors or four Intel RealSense D435 depth cameras~\cite{realsense}.
We trained our policy in PyTorch \cite{NEURIPS2019_9015} and deployed on the robot zero-shot without any fine-tuning.
We build a robot-centric 2.5D elevation map at 20 Hz by estimating the robot's pose and registering the point-cloud readings from the sensors accordingly.
The policy runs at 50 Hz and samples the heights from the latest elevation map, filling a randomly sampled value if no map information is available at a query location.

We developed an elevation mapping pipeline for fast terrain mapping on a \ac{GPU} to parallelize point-cloud processing.
We follow a similar approach to Fankhauser et al.~\cite{fankhauser2018probabilistic} to update the map in a Kalman-filter fashion and additionally perform drift compensation and ray casting to obtain a more consistent map. 
This fast mapping implementation was crucial to maintain fast processing rates and keep up with the fast locomotion speeds achieved by our controller.

\section{Acknowledgments}
\textbf{Funding} The project was funded, in part, by the Intel Network on Intelligent Systems, the Swiss National Science Foundation (SNF) through the National Centre of Competence in Research Robotics and project No.\ 188596, the European Research Council (ERC)
under the European Union’s Horizon 2020 research and innovation programme grant
agreement No.\ 852044, No.\ 780883 and No.\ 101016970. The work has been conducted as part of ANYmal Research, a community to advance legged robotics. \textbf{Author contributions} T.M. formulated the main idea of combining inputs from multiple modalities. J.L. and J.H designed and tested the initial setup. T.M. developed software and trained the controller. T.M. and L.W. set up the perception pipeline on the robot. T.M. conducted most of the indoor experiments. T.M., J.L., and L.W. conducted outdoor experiments. All authors refined ideas, contributed in the experiment design, analyzed the data, and wrote the paper. \textbf{Competing interests} The authors declare that they have no competing interests. \textbf{Data and materials availability} All data needed to evaluate the conclusions in the paper are present in the paper or the Supplementary Materials. 

\section*{Supplementary materials}
\makebox[1.8cm][l]{Section S1.} Nomenclature \\
\makebox[1.8cm][l]{Section S2.} Evaluating the importance of exteroception: \\
\makebox[1.8cm][l]{} simulation result \\
\makebox[1.8cm][l]{Section S3.} Training details \\
\makebox[1.8cm][l]{Section S4.} Terrain generation \\
\makebox[1.8cm][l]{Section S5.} Observation and action \\
\makebox[1.8cm][l]{Section S6.} Network architecture \\
\makebox[1.8cm][l]{Section S7.} Reward function \\
\makebox[1.8cm][l]{Section S8.} Height sample noise \\
\makebox[1.8cm][l]{Section S9.} Belief encoder evaluation \\
\makebox[1.8cm][l]{Figure S1.} Comparison of the presented controller to a \\ 
\makebox[1.8cm][l]{} proprioceptive baseline over random terrains\\
\makebox[1.8cm][l]{Figure S2.} Ablation analysis of the presented belief encoder\\
\makebox[1.8cm][l]{Table S1.} Hyperparameters for PPO \\
\makebox[1.8cm][l]{Table S2.} Hyperparameters for student training \\
\makebox[1.8cm][l]{Table S3.} Observations \\
\makebox[1.8cm][l]{Table S4.} Action difference between teacher and student under \\
\makebox[1.8cm][l]{} two exteroceptive noise conditions. \\
\makebox[1.8cm][l]{Movie S1.} Walking over stairs in different directions.\\ 
\makebox[1.8cm][l]{Movie S2.} Baseline comparison. \\ %
\makebox[1.8cm][l]{Movie S3.} Robustness evaluation. \\
\makebox[1.8cm][l]{Movie S4.} Slippery surface and soft obstacle.

\bibliographystyle{Science}

\bibliography{bibs}

\begin{thebibliography}{10}

\bibitem{raibert2008bigdog}
M.~Raibert, K.~Blankespoor, G.~Nelson, R.~Playter, Bigdog, the rough-terrain
  quadruped robot, {\it IFAC Proceedings Volumes\/}  10822--10825 (2008).

\bibitem{katz2019mini}
B.~Katz, J.~Di~Carlo, S.~Kim, Mini cheetah: A platform for pushing the limits
  of dynamic quadruped control, {\it 2019 International Conference on Robotics
  and Automation (ICRA)\/},  6295--6301 (IEEE, 2019).

\bibitem{hwangbo2019learning}
J.~Hwangbo, J.~Lee, A.~Dosovitskiy, D.~Bellicoso, V.~Tsounis, V.~Koltun,
  M.~Hutter, Learning agile and dynamic motor skills for legged robots, {\it
  Science Robotics\/} {\bf 4} (2019).

\bibitem{lee2020learning}
J.~Lee, J.~Hwangbo, L.~Wellhausen, V.~Koltun, M.~Hutter, Learning quadrupedal
  locomotion over challenging terrain, {\it Science Robotics\/} {\bf 5} (2020).

\bibitem{park2021jumping}
H.-W. Park, P.~M. Wensing, S.~Kim, Jumping over obstacles with {MIT} {Cheetah}
  2, {\it Robotics and Autonomous Systems\/} p. 103703 (2021).

\bibitem{spot}
{Boston Dynamics}, Spot, \url{https://www.bostondynamics.com/spot} (2021).
  [Online; accessed March-2021].

\bibitem{gehring2021anymal}
C.~Gehring, P.~Fankhauser, L.~Isler, R.~Diethelm, S.~Bachmann, M.~Potz,
  L.~Gerstenberg, M.~Hutter, {ANYmal} in the field: Solving industrial
  inspection of an offshore {HVDC} platform with a quadrupedal robot, {\it
  Field and Service Robotics\/},  247--260 (Springer, 2021).

\bibitem{agilityRobotics}
{Agility Robotics}, Robots, \url{https://www.agilityrobotics.com/robots}
  (2021). [Online; accessed June-2021].

\bibitem{unitree}
{Unitree Robotics}, A1, \url{https://www.unitree.com/products/a1/} (2021).
  [Online; accessed March-2021].

\bibitem{ghost}
{Ghost Robotics}, Vision 60, \url{https://www.ghostrobotics.io/} (2021).
  [Online; accessed June-2021].

\bibitem{matthis2018gaze}
J.~S. Matthis, J.~L. Yates, M.~M. Hayhoe, Gaze and the control of foot
  placement when walking in natural terrain, {\it Current Biology\/}
  1224--1233 (2018).

\bibitem{anymal}
{ANYbotics}, {ANYmal},
  \url{https://www.anybotics.com/anymal-autonomous-legged-robot/} (2021).
  [Online; accessed June-2021].

\bibitem{fankhauser2015kinect}
P.~Fankhauser, M.~Bloesch, D.~Rodriguez, R.~Kaestner, M.~Hutter, R.~Siegwart,
  Kinect v2 for mobile robot navigation: Evaluation and modeling, {\it 2015
  International Conference on Advanced Robotics (ICAR)\/},  388--394 (IEEE,
  2015).

\bibitem{ye2003new}
C.~Ye, J.~Borenstein, A new terrain mapping method for mobile robots obstacle
  negotiation, {\it Unmanned ground vehicle technology V\/},  52--62
  (International Society for Optics and Photonics, 2003).

\bibitem{belter2011rough}
D.~Belter, P.~Skrzypczyński, Rough terrain mapping and classification for
  foothold selection in a walking robot, {\it 2010 IEEE Safety Security and
  Rescue Robotics\/},  1--6 (2010).

\bibitem{fankhauser2014robot}
P.~Fankhauser, M.~Bloesch, C.~Gehring, M.~Hutter, R.~Siegwart.
\newblock Robot-centric elevation mapping with uncertainty estimates.
\newblock {\it Mobile Service Robotics\/} (World Scientific, 2014),  433--440.

\bibitem{fankhauser2018probabilistic}
P.~Fankhauser, M.~Bloesch, M.~Hutter, Probabilistic terrain mapping for mobile
  robots with uncertain localization, {\it IEEE Robotics and Automation
  Letters\/}  3019--3026 (2018).

\bibitem{zucker2010optimization}
M.~Zucker, J.~A. Bagnell, C.~G. Atkeson, J.~Kuffner, An optimization approach
  to rough terrain locomotion, {\it 2010 IEEE International Conference on
  Robotics and Automation\/},  3589--3595 (IEEE, 2010).

\bibitem{neuhaus2011comprehensive}
P.~D. Neuhaus, J.~E. Pratt, M.~J. Johnson, Comprehensive summary of the
  institute for human and machine cognition’s experience with {LittleDog},
  {\it The International Journal of Robotics Research\/}  216--235 (2011).

\bibitem{kolter2009stereo}
J.~Z. Kolter, Y.~Kim, A.~Y. Ng, Stereo vision and terrain modeling for
  quadruped robots, {\it 2009 IEEE International Conference on Robotics and
  Automation\/},  1557--1564 (IEEE, 2009).

\bibitem{havoutis2013onboard}
I.~Havoutis, J.~Ortiz, S.~Bazeille, V.~Barasuol, C.~Semini, D.~G. Caldwell,
  Onboard perception-based trotting and crawling with the hydraulic quadruped
  robot ({HyQ}), {\it 2013 IEEE/RSJ International Conference on Intelligent
  Robots and Systems\/},  6052--6057 (IEEE, 2013).

\bibitem{mastalli2017trajectory}
C.~Mastalli, M.~Focchi, I.~Havoutis, A.~Radulescu, S.~Calinon, J.~Buchli, D.~G.
  Caldwell, C.~Semini, Trajectory and foothold optimization using
  low-dimensional models for rough terrain locomotion, {\it 2017 IEEE
  International Conference on Robotics and Automation (ICRA)\/},  1096--1103
  (IEEE, 2017).

\bibitem{belter2016adaptive}
D.~Belter, P.~Łabęcki, P.~Skrzypczyński, Adaptive motion planning for
  autonomous rough terrain traversal with a walking robot, {\it Journal of
  Field Robotics\/}  337--370 (2016).

\bibitem{fankhauser2018robust}
P.~Fankhauser, M.~Bjelonic, C.~D. Bellicoso, T.~Miki, M.~Hutter, Robust
  rough-terrain locomotion with a quadrupedal robot, {\it 2018 IEEE
  International Conference on Robotics and Automation (ICRA)\/},  5761--5768
  (IEEE, 2018).

\bibitem{jenelten2020perceptive}
F.~Jenelten, T.~Miki, A.~E. Vijayan, M.~Bjelonic, M.~Hutter, Perceptive
  locomotion in rough terrain--online foothold optimization, {\it IEEE Robotics
  and Automation Letters\/}  5370--5376 (2020).

\bibitem{kim2020vision}
D.~Kim, D.~Carballo, J.~Di~Carlo, B.~Katz, G.~Bledt, B.~Lim, S.~Kim, Vision
  aided dynamic exploration of unstructured terrain with a small-scale
  quadruped robot, {\it 2020 IEEE International Conference on Robotics and
  Automation (ICRA)\/},  2464--2470 (IEEE, 2020).

\bibitem{magana2019fast}
O.~A. Villarreal{-}Maga{\~{n}}a, V.~Barasuol, M.~Camurri, M.~Focchi,
  L.~Franceschi, M.~Pontil, D.~G. Caldwell, C.~Semini, Fast and continuous
  foothold adaptation for dynamic locomotion through cnns, {\it IEEE Robotics
  and Automation Letters\/}  2140--2147 (2019).

\bibitem{atlasparkour2021}
{Boston Dynamics}, Atlas | partners in parkour,
  \url{https://youtu.be/tF4DML7FIWk} (2021). [Online; accessed September-2021].

\bibitem{peng2016terrain}
X.~B. Peng, G.~Berseth, M.~Van~de Panne, Terrain-adaptive locomotion skills
  using deep reinforcement learning, {\it ACM Transactions on Graphics (TOG)\/}
   1--12 (2016).

\bibitem{peng2017deeploco}
X.~B. Peng, G.~Berseth, K.~Yin, M.~Van De~Panne, Deeploco: Dynamic locomotion
  skills using hierarchical deep reinforcement learning, {\it ACM Transactions
  on Graphics (TOG)\/}  1--13 (2017).

\bibitem{2018-TOG-deepMimic}
X.~B. Peng, P.~Abbeel, S.~Levine, M.~van~de Panne, Deepmimic: Example-guided
  deep reinforcement learning of physics-based character skills, {\it ACM
  Trans. Graph.\/}  143:1--143:14 (2018).

\bibitem{xie2020allsteps}
Z.~Xie, H.~Y. Ling, N.~H. Kim, M.~van~de Panne, Allsteps: Curriculum-driven
  learning of stepping stone skills, {\it Computer Graphics Forum\/},  213--224
  (Wiley Online Library, 2020).

\bibitem{tsounis2020deepgait}
V.~Tsounis, M.~Alge, J.~Lee, F.~Farshidian, M.~Hutter, Deepgait: Planning and
  control of quadrupedal gaits using deep reinforcement learning, {\it IEEE
  Robotics and Automation Letters\/}  3699--3706 (2020).

\bibitem{tan2018sim}
J.~Tan, T.~Zhang, E.~Coumans, A.~Iscen, Y.~Bai, D.~Hafner, S.~Bohez,
  V.~Vanhoucke, Sim-to-real: Learning agile locomotion for quadruped robots,
  {\it Robotics: Science and Systems\/} (2018).

\bibitem{RoboImitationPeng20}
X.~B. Peng, E.~Coumans, T.~Zhang, T.-W.~E. Lee, J.~Tan, S.~Levine, Learning
  agile robotic locomotion skills by imitating animals, {\it Robotics: Science
  and Systems\/} (2020).

\bibitem{yang2020data}
Y.~Yang, K.~Caluwaerts, A.~Iscen, T.~Zhang, J.~Tan, V.~Sindhwani, Data
  efficient reinforcement learning for legged robots, {\it Conference on Robot
  Learning\/},  1--10 (PMLR, 2020).

\bibitem{pmlr-v100-xie20a}
Z.~Xie, P.~Clary, J.~Dao, P.~Morais, J.~Hurst, M.~van~de Panne, Learning
  locomotion skills for cassie: Iterative design and sim-to-real, {\it
  Proceedings of the Conference on Robot Learning\/}, L.~P. Kaelbling,
  D.~Kragic, K.~Sugiura, eds.,  317--329 (PMLR, 2020).

\bibitem{siekmann2021blind}
J.~Siekmann, K.~Green, J.~Warila, A.~Fern, J.~Hurst, Blind bipedal stair
  traversal via sim-to-real reinforcement learning, {\it Robotics: Science and
  Systems\/} (2021).

\bibitem{kumar2021rma}
A.~Kumar, Z.~Fu, D.~Pathak, J.~Malik, Rma: Rapid motor adaptation for legged
  robots, {\it Proceedings of Robotics: Science and Systems (RSS)\/} (2021).

\bibitem{yang2020}
C.~Yang, K.~Yuan, Q.~Zhu, W.~Yu, Z.~Li, Multi-expert learning of adaptive
  legged locomotion, {\it Science Robotics\/} p. eabb2174 (2020).

\bibitem{lee2019robust}
J.~Lee, J.~Hwangbo, M.~Hutter, Robust recovery controller for a quadrupedal
  robot using deep reinforcement learning, {\it arXiv preprint
  arXiv:1901.07517\/}  (2019).

\bibitem{gangapurwala2020rloc}
S.~Gangapurwala, M.~Geisert, R.~Orsolino, M.~Fallon, I.~Havoutis, {RLOC}:
  Terrain-aware legged locomotion using reinforcement learning and optimal
  control, {\it arXiv preprint arXiv:2012.03094\/}  (2020).

\bibitem{focchi2020heuristic}
M.~Focchi, R.~Orsolino, M.~Camurri, V.~Barasuol, C.~Mastalli, D.~G. Caldwell,
  C.~Semini.
\newblock Heuristic planning for rough terrain locomotion in presence of
  external disturbances and variable perception quality.
\newblock {\it Advances in Robotics Research: From Lab to Market\/} (Springer,
  2020),  165--209.

\bibitem{spotguide}
{Boston Dynamics}, Spot user guide release 2.0 version {A},
  \url{https://www.generationrobots.com/media/spot-boston-dynamics/spot-user-guide-r2.0-va.pdf
  } (2021). [Online; accessed June-2021].

\bibitem{chen2020learning}
D.~Chen, B.~Zhou, V.~Koltun, P.~Kr{\"a}henb{\"u}hl, Learning by cheating, {\it
  Conference on Robot Learning\/},  66--75 (PMLR, 2020).

\bibitem{komoot}
{Komoot}, Etzel kulm loop hike, \url{https://bit.ly/35bjfyE} (2021). [Online;
  accessed June-2021].

\bibitem{bloesch2013state}
M.~Bloesch, M.~Hutter, M.~A. Hoepflinger, S.~Leutenegger, C.~Gehring, C.~D.
  Remy, R.~Siegwart, State estimation for legged robots-consistent fusion of
  leg kinematics and imu, {\it Robotics\/}  17--24 (2013).

\bibitem{komoothelp}
{Komoot}, Komoot help guides,
  \url{https://d21buns5ku92am.cloudfront.net/67683/documents/40488-Komoot%20Guides%20English-4d1241.pdf}
  (2021). [Online; accessed December-2021].

\bibitem{coulter1992implementation}
R.~C. Coulter, Implementation of the pure pursuit path tracking algorithm, {\it
  Tech. rep.\/}, Carnegie-Mellon UNIV Pittsburgh PA Robotics INST (1992).

\bibitem{tranzatto2021cerberus}
M.~Tranzatto, F.~Mascarich, L.~Bernreiter, C.~Godinho, M.~Camurri, S.~M.~K.
  Khattak, T.~Dang, V.~Reijgwart, J.~Loeje, D.~Wisth, others, Cerberus:
  Autonomous legged and aerial robotic exploration in the tunnel and urban
  circuits of the darpa subterranean challenge, {\it Journal of Field
  Robotics\/}  (2021).

\bibitem{cerberus}
{Cerberus}, Team cerberus, \url{https://www.subt-cerberus.org/} (2021).
  [Online; accessed June-2021].

\bibitem{subtresult}
{DARPA}, Darpa subterranean challenge competition results finals,
  \url{https://www.subtchallenge.com/results.html} (2021). [Online; accessed
  November-2021].

\bibitem{subt}
{DARPA}, Darpa subterranean challenge competition rules final event,
  \url{https://www.subtchallenge.com} (2021). [Online; accessed June-2021].

\bibitem{mnih2013playing}
V.~Mnih, K.~Kavukcuoglu, D.~Silver, A.~Graves, I.~Antonoglou, D.~Wierstra,
  M.~Riedmiller, Playing atari with deep reinforcement learning, {\it Advances
  in Neural Information Processing Systems, Deep Learning Workshop\/}  (2013).

\bibitem{zhu2017improving}
P.~Zhu, X.~Li, P.~Poupart, G.~Miao, On improving deep reinforcement learning
  for pomdps, {\it arXiv preprint arXiv:1704.07978\/}  (2017).

\bibitem{vinyals2019grandmaster}
O.~Vinyals, I.~Babuschkin, W.~M. Czarnecki, M.~Mathieu, A.~Dudzik, J.~Chung,
  D.~H. Choi, R.~Powell, T.~Ewalds, P.~Georgiev, others, Grandmaster level in
  starcraft ii using multi-agent reinforcement learning, {\it Nature\/}
  350--354 (2019).

\bibitem{BaiTCN2018}
S.~Bai, J.~Z. Kolter, V.~Koltun, An empirical evaluation of generic
  convolutional and recurrent networks for sequence modeling, {\it
  arXiv:1803.01271\/}  (2018).

\bibitem{raisim}
J.~Hwangbo, J.~Lee, M.~Hutter, Per-contact iteration method for solving contact
  dynamics, {\it IEEE Robotics and Automation Letters\/}  895--902 (2018).

\bibitem{schulman2017proximal}
J.~Schulman, F.~Wolski, P.~Dhariwal, A.~Radford, O.~Klimov, Proximal policy
  optimization algorithms, {\it arXiv preprint arXiv:1707.06347\/}  (2017).

\bibitem{ross2011reduction}
S.~Ross, G.~Gordon, D.~Bagnell, A reduction of imitation learning and
  structured prediction to no-regret online learning, {\it Proceedings of the
  fourteenth international conference on artificial intelligence and
  statistics\/},  627--635 (JMLR Workshop and Conference Proceedings, 2011).

\bibitem{pmlr-v89-czarnecki19a}
W.~M. Czarnecki, R.~Pascanu, S.~Osindero, S.~Jayakumar, G.~Swirszcz,
  M.~Jaderberg, Distilling policy distillation, {\it Proceedings of Machine
  Learning Research\/}, K.~Chaudhuri, M.~Sugiyama, eds.,  1331--1340 (PMLR,
  2019).

\bibitem{cho2014learning}
K.~Cho, B.~Van~Merri{\"e}nboer, C.~Gulcehre, D.~Bahdanau, F.~Bougares,
  H.~Schwenk, Y.~Bengio, Learning phrase representations using rnn
  encoder-decoder for statistical machine translation, {\it Conference on
  Empirical Methods in Natural Language Processing (EMNLP)\/}, p. 1724–1734
  (2014).

\bibitem{hochreiter1997long}
S.~Hochreiter, J.~Schmidhuber, Long short-term memory, {\it Neural
  Computation\/}  1735--1780 (1997).

\bibitem{anzai2020deep}
T.~Anzai, K.~Takahashi, Deep gated multi-modal learning: In-hand object pose
  changes estimation using tactile and image data, {\it 2020 IEEE/RSJ
  International Conference on Intelligent Robots and Systems (IROS)\/},
  9361--9368 (IEEE, 2020).

\bibitem{kim2018robust}
J.~Kim, J.~Koh, Y.~Kim, J.~Choi, Y.~Hwang, J.~W. Choi, Robust deep multi-modal
  learning based on gated information fusion network, {\it Asian Conference on
  Computer Vision\/},  90--106 (Springer, 2018).

\bibitem{arevalo2017gated}
J.~Arevalo, T.~Solorio, M.~Montes-y G{\'o}mez, F.~A. Gonz{\'a}lez, Gated
  multimodal units for information fusion, {\it ICLR workshop\/}  (2017).

\bibitem{bpearl}
Rs-bpearl (2021, april), \url{https://www.robosense.ai/en/rslidar/RS-Bpearl}.

\bibitem{realsense}
Intel realsense (2021, april), \url{https://www.intelrealsense.com/}.

\bibitem{NEURIPS2019_9015}
A.~Paszke, S.~Gross, F.~Massa, A.~Lerer, J.~Bradbury, G.~Chanan, T.~Killeen,
  Z.~Lin, N.~Gimelshein, L.~Antiga, A.~Desmaison, A.~Kopf, E.~Yang, Z.~DeVito,
  M.~Raison, A.~Tejani, S.~Chilamkurthy, B.~Steiner, L.~Fang, J.~Bai,
  S.~Chintala.
\newblock Pytorch: An imperative style, high-performance deep learning library.
\newblock {\it Advances in Neural Information Processing Systems 32\/},
  H.~Wallach, H.~Larochelle, A.~Beygelzimer, F.~d\textquotesingle
  Alch\'{e}-Buc, E.~Fox, R.~Garnett, eds. (Curran Associates, Inc., 2019),
  8024--8035.

\bibitem{rslgym}
M.~Takahiro, L.~Joonho, M.~Yuntao, E.~Pascal, Rslgym, {\it GitHub repository\/}
   (2021).

\bibitem{kingma2014adam}
D.~P. Kingma, J.~Ba, Adam: A method for stochastic optimization, {\it ICLR\/}
  (2015).

\bibitem{lagae2010survey}
A.~Lagae, S.~Lefebvre, R.~Cook, T.~DeRose, G.~Drettakis, D.~S. Ebert, J.~P.
  Lewis, K.~Perlin, M.~Zwicker, A survey of procedural noise functions, {\it
  Computer Graphics Forum\/},  2579--2600 (Wiley Online Library, 2010).

\bibitem{maas2013rectifier}
A.~L. Maas, A.~Y. Hannun, A.~Y. Ng, Rectifier nonlinearities improve neural
  network acoustic models, {\it Proc. icml\/}, p.~3 (Citeseer, 2013).

\end{thebibliography}

\clearpage
\newpage

\setcounter{table}{0}
\makeatletter 
\renewcommand{\thetable}{S\@arabic\c@table}
\makeatother

\setcounter{figure}{0}
\makeatletter 
\renewcommand{\thefigure}{S\@arabic\c@figure}
\makeatother

\setcounter{algorithm}{0}
\makeatletter 
\renewcommand{\thealgorithm}{S\@arabic\c@algorithm}
\makeatother

\subsection*{S1. Nomenclature}
\makebox[1.3cm]{$s$} state\\
\makebox[1.3cm]{$o$} observation\\
\makebox[1.3cm]{$b$} belief state\\
\makebox[1.3cm]{$h$} hidden state\\
\makebox[1.3cm]{$l$} latent feature\\
\makebox[1.3cm]{$v$} linear velocity\\
\makebox[1.3cm]{$\omega$} angular velocity\\
\makebox[1.3cm]{$\tau$} joint torque\\
\makebox[1.3cm]{$q$} joint position\\
\makebox[1.3cm]{$\phi$} CPG phase\\
\makebox[1.3cm]{$\varDelta \phi_0$} CPG phase base frequency\\
\makebox[1.3cm]{$c_k$} curriculum factor\\
\makebox[1.3cm]{$c_{sk}$} student curriculum factor\\
\makebox[1.3cm]{$\mathcal{L}_{bc}$} behavior cloning loss\\
\makebox[1.3cm]{$\mathcal{L}_{re}$} reconstruction loss\\
\makebox[1.3cm]{$(\cdot)^p$} proprioceptive quantity\\
\makebox[1.3cm]{$(\cdot)^e$} exteroceptive quantity\\
\makebox[1.3cm]{$(\cdot)^{priv}$} privileged quantity\\
\makebox[1.3cm]{$(\cdot)^{target}$} target quantity\\
\makebox[1.3cm]{$(\cdot)_t$} quantity at time $t$\\
\makebox[1.3cm]{$\Tilde{(\cdot)}$} noisy quantity\\
\makebox[1.3cm]{$\dot{(\cdot)}$} first derivative\\
\makebox[1.3cm]{$\ddot{(\cdot)}$} second derivative\\
\makebox[1.3cm]{$g(\cdot)$} \ac{MLP} encoder \\
\makebox[1.3cm]{$\mathcal{N}()$} Normal distribution \\
\makebox[1.3cm]{$\odot$} Hadamard product\\
\makebox[1.3cm]{$\mathbf{p}(\cdot)$} foot trajectory function\\
\makebox[1.3cm]{$IK(\cdot)$} inverse kinematics function\\

\subsection*{S2. Evaluating the importance of exteroception: Additional experiments in simulation}

We compare the success rate over various stepped terrain and stairs in simulation to further evaluate the performance quantitatively.

The robot was given a fixed forward velocity command of 0.7 m/s for a duration of 10 seconds. 
We collected 300 trials to calculate the success rate, where we consider a trial a success if the robot can traverse 4 m without failure.
As shown in Figure 8A, 8B our controller significantly outperformed the baseline and can traverse a much wider range of terrain.

\subsection*{S3. Training details}

The control frequency of the policy was set to 50 Hz, and 250 trajectory time steps per environment are collected for one training iteration. We parallelized the simulation environment to perform rollouts with 1000 environments simultaneously.
We used our custom implementation of \ac{PPO}~\cite{schulman2017proximal} to train the teacher policy~\cite{rslgym}.
Observations are normalized using running mean and standard deviation before giving them to the policy network.
The curriculum factors were updated exponentially every training episode $c_{k+1} = c_k^d$, with convergence rate $d = 0.98$.
We use the Adam~\cite{kingma2014adam} optimizer with exponential learning rate decay.
The hyperparameters for \ac{PPO} are given in Table S1.

\begin{table}
\caption{Hyperparameters for \ac{PPO}.}
\centering
\begin{tabular}{l|ll}
\hline
learning rate               & 5.0 E-4 \\
learning rate decay gamma   & 0.9999  \\
discount factor             & 0.996 \\
learning epoch              & 2 \\
GAE-lambda                  & 0.95 \\
clip ratio                  & 0.2 \\
entropy coefficient         & 0.005 \\
batch size                  & 8300 \\

\end{tabular}
\end{table}

For student training, we performed rollouts with 300 environments and collected 400 timesteps of trajectory for one training iteration.
We start the student training without height sample noise and gradually increase the noise level through a student curriculum factor which linearly increases over training epochs.
We use flat terrain for the first 10 epochs, and then enable the adaptive curriculum for the terrain generation.
After 20 epochs, we increase the student curriculum factor $c_{sk}$ linearly  until we reach 100 epochs. Then, we keep  $c_{sk} = 1.0$.
We train the \ac{RNN} unit of the encoder with Truncated Backpropagation Through Time (TBPTT).
The ratio between behavior cloning loss and reconstruction loss is 0.5. Therefore the loss is set to $\mathcal{L}_{bc} + 0.5 \cdot \mathcal{L}_{re}$.
Hyperparameters for student training are given in Table S2.

\begin{table}
\caption{Hyperparameters for student training.}
\centering
\begin{tabular}{l|ll}
\hline
learning rate               & 5.0 E-4 \\
truncate step for TBPTT     & 10  \\
learning epoch              & 2 \\

\end{tabular}
\end{table}

\subsection*{S4. Terrain generation}
The terrain types are \textit{rough}, \textit{rough discrete}, \textit{large steps}, \textit{boxes}, \textit{grid steps}, \textit{step stairs}, and \textit{stairs}, as shown in Figure 6.1.
There are four types of stairs: \textit{standard stair}, \textit{open stair}, \textit{ledged stair}, and \textit{random stair}.
Each terrain type is parameterized by different terrain properties, which are randomized during training. 

The \textit{rough} terrain is parameterized by Perlin noise~\cite{lagae2010survey} and the \textit{rough discrete} and \textit{large steps} are created by quantizing it.
While \textit{rough discrete} terrain does not restrict the number of quantization levels, \textit{large steps} only allow for two height levels ($h \in [0, 0.4]~\unit{m}$).
For \textit{grid steps}, the parameters are mean step height ($ h \in [0.05, 0.4]~\unit{m}$) and step width ($d \in [0.2, 0.7]~\unit{m}$). Some examples of different \textit{grid steps} are shown in Figure 8A. Note that the parameter range shown in the figure is only for evaluation and different from the range used during training.
Parameters for \textit{stairs} contain step depth ($ d \in [0.25, 1.0] ~\unit{m}$) and height ($h \in [0.01, 0.22]~\unit{m}$).
The height and depth values for \textit{random stair} were set at each according to a ratio $\epsilon \sim \mathcal{N}(1.0, 0.2)$, such that $\hat{x} = x \cdot \epsilon$, where $x$ is the given depth or height parameter.
Examples of different stairs are shown in Figure 8B.
The \textit{boxes} terrain consists of multiple boxes with maximum height 0.25~\unit{m} lying in a random position with random yaw angles.

\subsection*{S5. Observation and action}
The observation vectors are defined in Table S3. Proprioceptive input includes command, joint, and body information, as well as leg phase information. The \ac{CPG}'s phase information consists of $\varDelta \phi_l$, $\cos\phi_l$, $\sin\phi_l$, and base frequency for each leg $l$. 
For exteroception, we use height samples around each foot instead of the local elevation map.
The circular sampling pattern comprises \{6, 8, 10, 12, 16\} points around each foot, with radii \{0.08, 0.16, 0.26, 0.36, 0.48\} m, respectively.

\begin{table}[htbp]
\caption{Observations. Proprioception is used for both teacher and student training. Exteroception is given in the form of height samples. The privileged information is used only for teacher training.}
\begin{tabular}{l|ll}
\hline
Observation type & Input                   & Dim. \\ \hline
Proprioception   & command                 & 3             \\
                 & body orientation        & 3             \\
                 & body velocity           & 6             \\
                 & joint position          & 12            \\
                 & joint velocity          & 12            \\
                 & joint position history (3 time steps)  & 36            \\
                 & joint velocity history (2 time steps) & 24            \\
                 & joint target history   (2 time steps) & 24            \\
                 & CPG phase information & 13            \\ \hline
Exteroception    & height samples             & 208           \\ \hline
Privileged info. & contact states              & 4             \\
                 & contact forces              & 12            \\
                 & contact normals             & 12            \\
                 & friction coefficients       & 4             \\
                 & thigh and shank contact     & 8             \\
                 & external forces and torques & 6             \\
                 & airtime                     & 4             \\ \hline
\end{tabular}
\label{tab:proprioceptive}
\end{table}

The action is defined as $\langle \varDelta \phi_l, \varDelta q_i \rangle$, where $\varDelta \phi_l$ and $\varDelta q_i$ refer to the phase offset per leg ($l \in \{ legs \}$) and the residual joint position target ($i \in \{1, \cdots, 12\}$), respectively.
We have a nominal foot trajectory $\mathbf{p}(\phi): \mathbb{R} \longrightarrow \mathbb{R}^3$ that maps each $\phi_l$ to a target foot position, which generates periodic stepping motion as $\phi$ cycles within $[0, 2\pi)$.
From the action, the joint position target for a leg $l$ is defined as $q_{i\in{l}}^{target} = IK(\mathbf{p}(\phi_l + \varDelta \phi_l + \varDelta \phi_0)) + \varDelta q_{i\in{l}}$, using analytic inverse kinematics $IK(\cdot)$ and base phase frequency $\varDelta \phi_0$.
The nominal foot trajectory is defined as follows.

If the phase is in swing-up ($0 \leq \phi_l \leq \pi / 2$),
\begin{eqnarray*}
        \mathbf{p}_l(\phi_l) &= \langle x_l^n, y_l^n, z_l^n + 0.2 \cdot (-2t_l^3 + 3t_l^2) \rangle, \\
        & \text{where} \quad  t_l = 2 / \pi \cdot \phi_l.
\end{eqnarray*}
$\{x,y,z\}_l^n$ is the nominal foot position at the default stance configuration.
The cubic Hermite spline connects $z = z_l^n $ at $\phi_l = 0$ and $z = z_l^n  + 0.2$ at $\phi_l = \pi/2$.

In the swing-down phase ($\pi / 2 < \phi_l \leq \pi$), the foot height is computed as
\begin{eqnarray*}
        \mathbf{p}_l(\phi_l) &= \langle x_l^n, y_l^n, z_l^n + 0.2 \cdot (2t_l^3 - 3t_l^2 + 1) \rangle, \\
        & \text{where} \quad  t_l = 2 / \pi \cdot \phi_l - 1,
\end{eqnarray*}
which is symmetric to the previous function.

During the stance phase ($\pi < \phi_l \leq 2\pi$), $\mathbf{p}_l(\phi_l) = \langle x_l^n, y_l^n, z_l^n \rangle$.

\subsection*{S6. Network architecture}
The policy network is composed of multiple \acp{MLP}.
The height samples are first encoded into a $24 \times 4 = 96$ dimensional latent vector, and the privileged information is encoded into a 24 dimensional latent vector using MLP-based encoders ($g_e$, $g_p$). Each encoder has two hidden layers with \{80, 60\} and \{64, 32\} hidden units respectively. The height samples are first fed into the encoder separately for each foot and then concatenated into one feature vector.
Then these features are concatenated with proprioceptive observations and fed into another MLP with three hidden layers \{256, 160, 128\}. The activation function for all \acp{MLP} is LeakyReLU \cite{maas2013rectifier}.

We use a \ac{GRU} with an exteroceptive gate for the belief encoder (Figure 7C).
The \ac{GRU} consists of 2 stacked layers with 50 hidden units each.
The belief encoder and exteroceptive gate $g_b$, $g_a$ are used to calculate $96 + 24 = 120 $ dimensional belief state $b_t$ and $96$ dimensional attention vector $\alpha$. Each encoder has two hidden layers with \{64, 64\} and \{64, 64\} hidden units each. The filtered exteroceptive information $l_t^e \odot  \alpha $ is added to $g_b(b_t')$, with zero-padding to match the dimensionality.

\subsection*{S7. Reward function}

The reward function is defined as
$r = 0.75(r_{lv} + r_{av} + r_{lvo}) + r_b + 0.003r_{fc} + 0.1r_{co} + 0.001r_{j} + 0.08r_{jc} +     0.003 r_s+ 1.0 \cdot 10^{-6} r_{\tau} + 0.003r_{slip}$.
The individual terms are defined as follows.

\begin{itemize}
    \item Linear Velocity Reward ($r_{lv}$): This term encourages the policy to follow a desired horizontal velocity (velocity in $xy$ plane) command:
    \begin{equation*}
    r_{lv} = 
    \begin{cases}
    \exp(-|\bm{v}|^2), & \text{if } |\bm{v}_{des}| = 0 \\
    1.0, & \text{else if } \bm{v}_{des} \cdot \bm{v} > |\bm{v}_{des}|\\
    \exp(-(\bm{v}_{des} \cdot \bm{v} - |\bm{v}_{des}|)^2),& \text{otherwise}
     \end{cases}
\end{equation*}
where $\bm{v}_{des} \in \mathbb{R}^2$ is the desired horizontal velocity and $\bm{v}\in \mathbb{R}^2$ is the current body velocity with respect to the body frame.

   \item Angular Velocity Reward ($r_{av}$): This term encourages the policy to follow a desired yaw velocity command:
    \begin{equation*}
    r_{av} = 
    \begin{cases}
    \exp(-\omega_z^2), & \text{if } \omega_{des} = 0 \\

    1.0, & \text{else if } \omega_{des} \cdot \omega_z > \omega_{des}\\
    \exp(-(\omega_{des} \cdot \omega_z - \omega_{des})^2),& \text{otherwise}
     \end{cases}
\end{equation*}
     where $\omega_{des}$ is the desired yaw velocity and $\omega_z$ is the current yaw velocity with respect to the body frame.
    
    \item Linear Orthogonal Velocity Reward ($r_{lvo}$): This term penalizes the velocity orthogonal to the target direction:
    \begin{equation*}
    r_{lvo} = 
    \exp(- 3.0|\bm{v}_o|^2),
\end{equation*}
where $\bm{v_o} = \bm{v} - (\bm{v}_{des} \cdot \bm{v}) \bm{v}_{des}$.

    \item Body motion Reward ($r_{b}$): This term penalizes the body velocity in directions not part of the command:
    \begin{equation*}
    r_{bm} = -1.25 v_z^2 - 0.4 |\omega_x| - 0.4 |\omega_y|.
\end{equation*}

    \item Foot Clearance Reward ($r_{fc}$):
    When a leg is in swing phase, i.e., $\phi_i \in [0, \pi)$, the robot should lift the corresponding foot higher than its surroundings. However, to prevent the robot from manifesting unnecessarily high foot clearance, we give a penalty reward $r_{fcl}$ to regularize the leg trajectory.
    $H_{sample,l}$ is the set of sampled heights around the $l$-th foot.
    Then, the clearance cost is defined as
    \begin{eqnarray*}
        r_{fcl} &=& \begin{cases}
   -1.0, & \text{if } \max ( H_{sample, l}) < -0.2 \\
   0.0 & \text{otherwise}
     \end{cases} \\
    r_{fc} &=& \sum_{l=1}^4 r_{fcl}
    \end{eqnarray*}
    Note that height samples are sampled with respect to the foot height, therefore -0.2 means the terrain is 0.2 m lower than the foot; ergo, the foot is 0.2 m higher than the sampled terrain height.
    
    \item Shank and Knee Collision Reward ($r_{co}$): We want to penalize undesirable contact between the terrain and robot parts other than the foot, to avoid hardware damage: 
    \begin{equation*}
    r_{co} = \begin{cases}
        -c_k, & \text{if shank or knee is in collision}\\
        0.0 & \text{otherwise}
     \end{cases} \\
    \end{equation*}
    where $c_k$ is the curriculum factor that increases monotonically and converges to 1.
   
   \item Joint Motion Reward ($r_{j}$): This term penalizes joint velocity and acceleration to avoid vibrations:
   \begin{equation*}
        r_{s} = - c_k\sum_{i=1}^{12} (
        0.01\dot{q_i}^2 + \ddot{q_{i}}^2
        ),
    \end{equation*}
    where $\dot{q_i}$ and $\ddot{q_{i}}$ are the joint velocity and acceleration, respectively.
    
    \item Joint Constraint Reward ($r_{jc}$): This term introduces a soft constraint in the joint space. To avoid the knee joint flipping in the opposite direction, we give a penalty for exceeding a threshold:
   \begin{eqnarray*}
        r_{jc,i} &=& \begin{cases}
        -(q_i - q_{i, th})^2, & \text{if } q_i > q_{i, th}\\
        0.0 & \text{otherwise}
     \end{cases} \\
        r_{jc} &=& \sum_{i=1}^{12} r_{jc,i}
    \end{eqnarray*}
    where $q_{i, th}$ is a threshold value for the $i$th joint.
    We only set thresholds for the knee joint.
    \item Target Smoothness Reward ($r_{s}$): 
    The magnitude of the first and second order finite difference derivatives of the target foot positions are penalized such that the generated foot trajectories become smoother:
     \begin{equation*}
        r_{s} = - c_k\sum_{i=1}^{12} (
        (q_{i, t}^{des} - q_{i, t-1}^{des})^2 + 
        (q_{i, t}^{des} - 2q_{i, t-1}^{des} + q_{i, t-2}^{des})^2
        ),
    \end{equation*}
    where $q^{des}_{i,t}$ is the joint target position of joint $i$ at time step $t$.
    
    \item Torque Reward ($r_{\tau}$): We penalize joint torques to reduce energy consumption ($\tau \propto \text{electric current}$):
    \begin{equation*}
          r_{\tau} = - c_k\sum_{i=1}^{12}\tau_i^2 ,
    \end{equation*}
    where $\tau_i$ is the $i$th joint's torque calculated as output by the actuator network.
    
    \item Slip Reward ($r_{slip}$): We penalize the foot velocity if the foot is in contact with the ground to reduce slippage:
    \begin{equation*}
          r_{slip} = - c_k\sum_{l\in \{ \text{foot in contact} \} }v_{f,l}^2 ,
    \end{equation*}
    where $v_{f,l}$ is the velocity of $l$th foot in contact with the ground.
\end{itemize}

\subsection*{S8. Height sample noise}
During student training, we randomize the height samples drawn around each foot (Figure 7A).
We perturbed the position of each sample and add noise to the measured height value as follows.
\begin{eqnarray*}
    x_p = r_p \cos(\theta_p) + \epsilon_{px} + \epsilon_{fx} + w_x\\\
    y_p = r_p \sin(\theta_p) + \epsilon_{py} + \epsilon_{fy} + w_y \\
    h_p = h(x_p, y_p) + \epsilon_{pz} + \epsilon_{fz} + w_z + \epsilon_{outlier}
\end{eqnarray*}
where $h(x_p, y_p)$ refers to the terrain height at position $(x_p, y_p)$. 
$r_p$ is the radial distance of the point $p$ and $\theta_p$ is the azimuthal angle of $p$ in polar coordinates around the foot.
$\epsilon_{px},\epsilon_{py},\epsilon_{pz} $ represents the noise that is sampled for each individual point every time step. 
$\epsilon_{fx},\epsilon_{fy},\epsilon_{fz} $ represents the noise that is sampled for each foot every time step.
$w_{x},w_{y},w_{z} $ represents the noise that is sampled for each foot per episode.
$\epsilon_{outlier}$ is a large noise intermittently added to simulate outliers.

Each noise is sampled from the normal distribution using the parameter $z$.
$\epsilon_{px}, \epsilon_{py} \sim\mathcal{N}(0, z_0)$, $\epsilon_{pz} \sim\mathcal{N}(0, z_1)$, 
$\epsilon_{fx}, \epsilon_{fy} \sim\mathcal{N}(0, z_2)$, $\epsilon_{fz} \sim\mathcal{N}(0, z_3)$, 
$\epsilon_{outlier} \sim\mathcal{N}(0, z_4)$ with probability $p = z_5$, 
$w_x, w_y \sim\mathcal{N}(0, z_6)$, $w_z \sim\mathcal{N}(0, z_7)$.

We defined three conditions for the student training; \textit{nominal}, \textit{offset}, \textit{noisy}.
Each parameter $z$ is defined as follows.
\begin{eqnarray}
     z_{nominal} &=& \langle 0.004, 0.005, 0.01, 0.04, 0.03, 0.05, 0.1 \rangle \\
     z_{offset} &=& \langle 0.004, 0.005, 0.01, 0.1c_{sk}, 0.1c_{sk}, 0.02, 0.1 \rangle \\
     z_{noisy} &=& \langle 0.004, 0.1c_{sk}, 0.1c_{sk}, 0.3c_{sk}, 0.3c_{sk}, 0.3c_{sk}, 0.1 \rangle 
\end{eqnarray}
where $c_{sk}$ is the student curriculum factor which linearly increases over training episodes.
We randomly picked one of the conditions at the beginning and in the middle of a trajectory. The probabilities are 60\%, 30\% and 10\%, respectively.

\subsection*{S9. Ablation study of attention gate in belief encoder}
We evaluated the effect of the exteroceptive gate by comparing the performance of the belief encoder with and without the gate. 
For this purpose, we trained four student policies using different belief encoders: "GRU gate", "GRU no gate", "MLP gate" and "MLP no gate". "GRU gate" uses the proposed exteroceptive gate while "GRU no gate" does not use it. "MLP" uses feed forward network instead of the recurrent unit.
Figure S2A shows the learning curve of the student training using four different architectures. The result shows that using a recurrent unit improves the performance. MLP failed to reconstruct the privileged information. Moreover, the exteroceptive gate constantly improves the performance for both GRU and MLP architectures. Note that in the beginning of the training, we started without exteroceptive noise and terrain curriculum, and increased them gradually. This effect can be seen as a steep increase of losses and decrease of reward in the beginning.

To evaluate the learned model, we collected 300 time steps with 100 different terrain parameters for each terrain type with two noise conditions: \textit{small} and \textit{large}.
Each noise parameter $z$ are defined as follows,
\begin{eqnarray}
     z_{small} &=& \langle 0.004, 0.005, 0.04, 0.04, 0.04, 0.01, 0.1 \rangle \\
     z_{large} &=& \langle 0.004, 0.3, 0.2, 0.1, 0.1, 0.03, 0.1 \rangle 
\end{eqnarray}
Then we calculated the squared distance between student action and teacher action, as well as decoded height samples and ground-truth height samples.
As shown in Table S4, S5, the gated encoder outperformed the non-gated encoder for both noise cases. 
The encoder utilizes the exteroceptive input through the skip connection when the exteroception is reliable.
When the height samples contain large noise, the exteroception does not provide reliable information. 
In this case, the gated structure and non-gated structure perform similarly (Table S4, S5). 
This indicates that the gated structure facilitates the use of exteroceptive information when it is reliable but does not sacrifice robustness when it becomes unreliable. 

To further evaluate the policies' performance, a step traversal success rate were compared against each policy. The robot was initialized in front of various height of step and given a constant velocity command (0.8 m/s) towards the step. We collected 100 trials for each height of the step and showed the success rate in Figure S2B. The result shows that "GRU gate" performs the best for both small noise and large noise case. As seen in the small noise case, the difference between "GRU gate" and "GRU no gate" is bigger than the large noise case. This supports that the gated structure can utilize exteroceptive information more when it is reliable.

\begin{table*}
\caption{
Action difference between teacher and student under different noise conditions. 
The quantities are presented as empirical means with standard deviations.
The belief encoder with the exteroceptive gate exhibits smaller action difference for all types of terrain when the noise is small. 
When the exteroception is unreliable (large noise), they perform similarly; 
this indicates that the gate blocks the skip connection such that our encoder becomes similar to the proprioceptive model in this condition.
}
\centering
\begin{tabular}{l|ll||ll}
        & \multicolumn{2}{c||}{Small exteroceptive noise} & \multicolumn{2}{c}{Large exteroceptive noise} \\ \hline
terrain              & ours          & without gate   & ours               & without gate        \\ \hline
rough                &\bfseries 0.690$\pm$0.40  &0.746$\pm$0.40      &\bfseries 0.879$\pm$0.46  &0.997$\pm$0.44           \\
rough discrete       &\bfseries 0.787$\pm$0.45  &0.857$\pm$0.54      &\bfseries 0.878$\pm$0.53  &0.964$\pm$0.55           \\
step stair           &\bfseries 0.652$\pm$0.39  &0.687$\pm$0.43      &\bfseries 0.975$\pm$0.49  &1.043$\pm$0.50           \\
large step           &\bfseries 0.719$\pm$0.40  &0.855$\pm$0.43      &\bfseries 1.142$\pm$0.55  &1.225$\pm$0.54           \\
grid steps           &\bfseries 1.444$\pm$0.56  &1.674$\pm$0.58      &2.218$\pm$0.70            &\bfseries 2.212$\pm$0.70 \\
standard stair       &\bfseries 0.854$\pm$0.67  &0.961$\pm$0.72      &\bfseries 1.387$\pm$0.59  &1.438$\pm$0.56           \\
open stair           &\bfseries 0.842$\pm$0.61  &0.938$\pm$0.65      &\bfseries 1.356$\pm$0.55  &1.428$\pm$0.53           \\
ledged stair         &\bfseries 0.819$\pm$0.39  &0.929$\pm$0.42      &\bfseries 1.373$\pm$0.53  &1.416$\pm$0.54           \\
boxes                &\bfseries 0.928$\pm$0.53  &1.123$\pm$0.56      &\bfseries 1.614$\pm$0.64  &1.683$\pm$0.68           \\
random stair         &\bfseries 0.872$\pm$0.45  &0.956$\pm$0.46      &\bfseries 1.489$\pm$0.59  &1.526$\pm$0.58
\end{tabular}                                                        
\end{table*}

\begin{table*}
\caption{Reconstruction error of height samples under different noise conditions.  The quantities are presented as empirical means with standard deviations. The belief encoder with the exteroceptive gate had smaller reconstruction error for all types of terrain. This shows the effectiveness of the gated skip connection when the exteroception is reliable.
When the noise is large, the gated encoder also performed better than the non-gated encoder, although the difference was smaller than in the small-noise setting.
}
\centering
\begin{tabular}{l|ll||ll}
        & \multicolumn{2}{c||}{Small exteroceptive noise} & \multicolumn{2}{c}{Large exteroceptive noise} \\ \hline
terrain              & ours               & without gate     & ours       & without gate                \\ \hline
rough                &\bfseries 1.21E-03$\pm$2.8E-04&1.36E-03$\pm$6.1E-04  &  \bfseries 1.03E-03$\pm$2.3E-04&1.17E-03$\pm$5.9E-04             \\
rough discrete       &\bfseries 9.99E-04$\pm$3.3E-04&1.03E-03$\pm$3.9E-04  &  \bfseries 1.02E-03$\pm$3.5E-04&1.05E-03$\pm$3.5E-04             \\
step stair           &\bfseries 1.13E-03$\pm$4.4E-04&1.31E-03$\pm$4.7E-04  &  \bfseries 1.41E-03$\pm$4.3E-04&1.48E-03$\pm$4.6E-04             \\
large step           &\bfseries 1.37E-03$\pm$8.0E-04&2.03E-03$\pm$1.0E-03  &  \bfseries 1.95E-03$\pm$8.2E-04& \bfseries 1.95E-03$\pm$7.8E-04  \\
grid steps           &\bfseries 3.05E-03$\pm$4.1E-04&4.77E-03$\pm$7.4E-04  &  \bfseries 4.17E-03$\pm$5.0E-04&4.39E-03$\pm$5.1E-04             \\
standard stair       &\bfseries 2.59E-03$\pm$2.2E-03&3.11E-03$\pm$2.2E-03  &  \bfseries 2.68E-03$\pm$1.6E-03&2.69E-03$\pm$1.5E-03             \\
open stair           &\bfseries 2.61E-03$\pm$2.3E-03&3.06E-03$\pm$2.0E-03  &  \bfseries 2.63E-03$\pm$1.2E-03&2.64E-03$\pm$1.1E-03             \\
ledged stair         &\bfseries 2.53E-03$\pm$1.7E-03&3.03E-03$\pm$1.5E-03  &  \bfseries 2.62E-03$\pm$1.2E-03&2.63E-03$\pm$1.1E-03             \\
boxes                &\bfseries 2.13E-03$\pm$1.4E-03&3.38E-03$\pm$1.5E-03  &  \bfseries 3.00E-03$\pm$1.0E-03&3.09E-03$\pm$1.2E-03             \\
random stair         &\bfseries 2.31E-03$\pm$9.1E-04&2.89E-03$\pm$8.2E-04  &  \bfseries 2.72E-03$\pm$7.9E-04&2.74E-03$\pm$8.0E-04             
\end{tabular}                                                        
\end{table*}

\clearpage

\renewcommand{\thefigure}{S\arabic{figure}}
\setcounter{figure}{0}

\begin{figure*}
   \centering
      \includegraphics[width=1.0\textwidth]{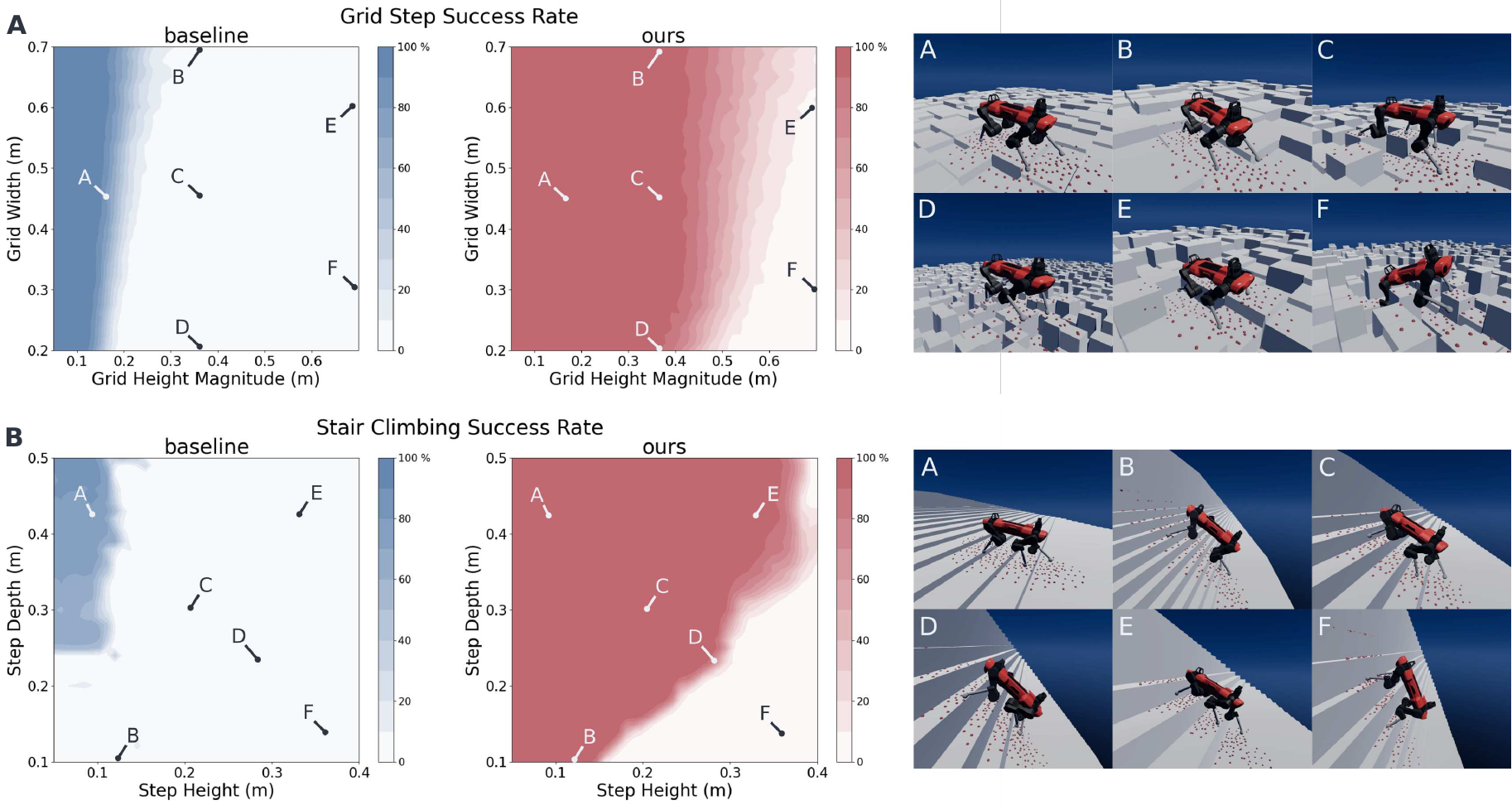}
    \caption{Comparison of the presented controller to a proprioceptive baseline~\cite{lee2020learning} over random terrain. We collected 300 trials with a fixed velocity command over $41 \times 41$ different terrain parameter combinations and compared success rates. Our controller was able to traverse a much wider range of terrain profiles on both grid steps (A) and stairs (B).}
    \label{fig:speed}
\end{figure*}

\begin{figure*}
   \centering
      \includegraphics[width=1.0\textwidth]{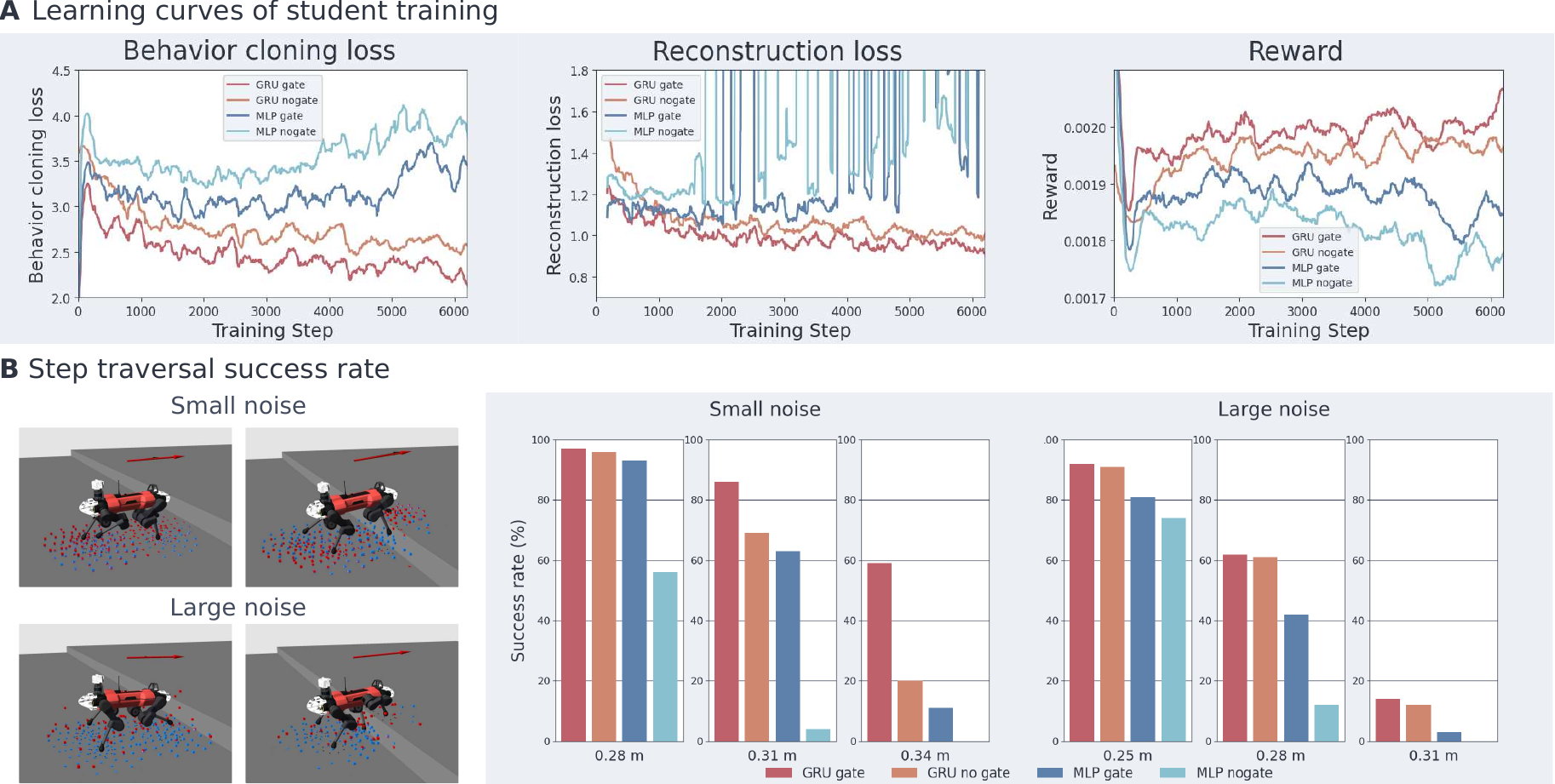}
    \caption{Ablation analysis of the presented belief encoder. We compared GRU gate, GRU no gate, MLP gate and MLP no gate. MLP setting uses MLP instead of GRU as its encoder. Gate setting uses proposed attention gate while no gate setting exclude it.(A) Learning curve of the student policy training. GRU worked better than MLP in all cases. Attention gate worked better than without attention for both GRU and MLP. The increase of the losses and decrease of reward in the beginning is due to the curriculum. (B) Step traversal success rate tested in small noise and large noise cases. The robot is initialized with random joint configuration and initial velocity and given a constant command towards the step. If the robot traversed the step with both front and hind legs it is considered as success. 100 trials were conducted.}
    \label{fig:speed}
\end{figure*}

\end{document}